\documentclass[]{jingdong}
\usepackage[toc,page,header]{appendix}
\usepackage{latexsym}
\usepackage[T1]{fontenc}
\usepackage[utf8]{inputenc}
\usepackage{microtype}
\usepackage{inconsolata}
\usepackage{graphicx}
\usepackage{hyperref}       
\usepackage{url}            
\usepackage{booktabs}       
\usepackage{amsfonts}       
\usepackage{nicefrac}       
\usepackage{stackengine}
\usepackage{microtype}      
\usepackage{colortbl}
\usepackage{xcolor}
\usepackage{amsmath}
\usepackage{amssymb}
\usepackage{amsthm}
\usepackage{mathrsfs}
\usepackage{pifont}
\usepackage{MnSymbol}
\usepackage{balance}
\usepackage{enumitem}
\usepackage{listings}
\usepackage{xcolor}
\usepackage{natbib}
\usepackage{multicol}

\AtBeginDocument{%
  \providecommand\BibTeX{{%
    \normalfont B\kern-0.5em{\scshape i\kern-0.25em b}\kern-0.8em\TeX}}}

\makeatletter
\DeclareRobustCommand\onedot{\futurelet\@let@token\@onedot}
\def\@onedot{\ifx\@let@token.\else.\null\fi}

\usepackage{setspace}
\usepackage{mathtools}

\usepackage{multirow,booktabs}
\usepackage{subcaption}

\newcommand{\owo}[1]{\textsc{OAgents}}

\definecolor{lightgreen}{RGB}{144, 238, 144} 
\definecolor{lightred}{RGB}{255, 105, 97}

\newtcolorbox{promptbox}[2][Prompt]{
colback=black!5!white,
arc=5pt, 
boxrule=0.5pt,
fonttitle=\bfseries,
title=#1, 
before upper={\small}, fontupper=\fontfamily{ptm}\selectfont,
colframe=#2, 
}
\definecolor{ogreen}{RGB}{34, 139, 34}

\usepackage{cleveref}
\theoremstyle{plain}

\theoremstyle{definition}

\theoremstyle{remark}
\usepackage{subcaption}

\usepackage[textsize=tiny]{todonotes}
\usepackage{fontawesome}

\usepackage{xcolor}
\usepackage{graphicx}
\usepackage[edges]{forest}
\usepackage{amsmath}

\usepackage{hyperref}
\usepackage{url}
\usepackage{graphicx}  
\usepackage{booktabs}
\usepackage{multirow}
\usepackage{xcolor}
\usepackage{pifont}
\usepackage{subcaption}
\usepackage[table]{xcolor}
\usepackage{xcolor}
\usepackage{tikz}
\usepackage{tcolorbox}
\tcbuselibrary{skins, breakable} 

\newcommand{\cmark}{\hspace{1.5pt}\ding{51}}
\newcommand{\xmark}{\hspace{1.5pt}\ding{55}} 

\definecolor{token11}{RGB}{254, 230, 149}
\definecolor{token12}{RGB}{242, 186, 2}
\newcommand{\tokentypeone}{%
  \tikz[baseline=-0.7ex]{
    \draw[
      fill=token11!50,
      draw=token12,
      line width=1.5pt
    ] (0,0) circle (1ex);
  }\hspace{3pt}%
}

\definecolor{token21}{RGB}{217, 217, 217}
\definecolor{token22}{RGB}{128, 128, 128}
\newcommand{\tokentypetwo}{%
  \tikz[baseline=-0.7ex]{
    \draw[
      fill=token21!50,
      draw=token22,
      line width=1.5pt
    ] (0,0) circle (1ex);
  }\hspace{3pt}%
}

\definecolor{token31}{RGB}{200, 229, 179}
\definecolor{token32}{RGB}{117, 189, 6}
\newcommand{\tokentypethree}{%
  \tikz[baseline=-0.7ex]{
    \draw[
      fill=token31!50,
      draw=token32,
      line width=1.5pt
    ] (0,0) circle (1ex);
  }\hspace{3pt}%
}

\definecolor{token41}{RGB}{182, 189, 234}
\definecolor{token42}{RGB}{72, 116, 203}
\newcommand{\tokentypefour}{%
  \tikz[baseline=-0.7ex]{
    \draw[
      fill=token41!50,
      draw=token42,
      line width=1.5pt
    ] (0,0) circle (1ex);
  }\hspace{3pt}%
}

\definecolor{mygreen}{RGB}{200, 229, 179}
\definecolor{myblue}{RGB}{182, 199, 234}

\newtcolorbox[]{takeaway}[1][]{%
  enhanced,
  breakable,
  colback=token31!70,          %
  colframe=token32!80,        %
  boxrule=1.5pt,
  arc=4pt,                  %
  left=2mm,right=2mm,top=2.5mm,bottom=2mm,
  before skip=10pt, after skip=10pt,
  colbacktitle=black!85,    %
  coltitle=white,           %
  fonttitle=\bfseries,
  attach boxed title to top center={yshift=-2.5mm},
  boxed title style={
    enhanced,
    arc=3pt,
    top=0.5mm, bottom=0.5mm, left=3mm, right=3mm,
    boxrule=0pt,           %
    interior engine=empty, %
  },
  #1                        %
}

\title{DOPD: Dual On-policy Distillation}
\author[1,2,4]{Xinlei Yu}
\author[4]{Gen Li}
\author[4]{Qingyi Si}
\author[1]{Guibin Zhang}
\author[4]{Yuqi Xu}
\author[4]{Congcong Wang}
\author[4]{Shuai Dong}
\author[4]{Kaiwen Tuo}
\author[4]{Xiangyu Zeng}
\author[2]{Kaituo Feng}
\author[2]{Qunzhong Wang}
\author[3]{Yang Shi}
\author[1,\text{\faEnvelopeO}]{Xiaobin Hu}
\author[2]{Xiangyu Yue}
\author[4]{Jiaqi Wang}
\author[1]{Shuicheng Yan}

\affiliation{
1 NUS \quad 2 MMLab, CUHK \quad 3 PKU \quad 4 Explore Academy, JD
}

\vspace{-20pt}
\abstract{

On-policy distillation (OPD) offers superior capacity transfer by supervising student-sampled trajectories with dense token-level signals. 
To furnish high-quality supervision sources and thereby elevate the performance frontier of distillation, an intuitive direction is to infuse privileged information to either teacher or student itself. 
However, this additional input induces a potential failure mode we dub \textbf{privilege illusion}: a pattern that conflates the transferable capability gap that students are meant to close, and the information asymmetry gap that can only be mimicked but never replicated.
This issue is further amplified by the inherent non-uniformity of token-level supervision, where only a small subset of tokens carries pivotal capability-bearing signals.
To this end, we propose \textbf{DOPD}, an advantage-aware dual distillation paradigm that dynamically routes token-level supervision between privileged teacher and privileged student policies based on their advantage gap and relative probabilities.
Each token receives supervision of different strength, objective, and strategy from either teacher or student itself, which transfers credible capability while simultaneously receiving auxiliary signals, to alleviate privilege illusion.
Extensive experiments on both large language model (LLM) and vision-language model (VLM) settings demonstrate that DOPD consistently outperforms Vanilla OPD and other counterparts. Further results on stability, robustness, continual learning, and out-of-distribution tasks validate its superiority.

}

\vspace{-20pt}

\begin{document}
\maketitle

\vspace{-20pt}

\begin{figure}[h]
    \centering
    \includegraphics[width=0.92\linewidth]{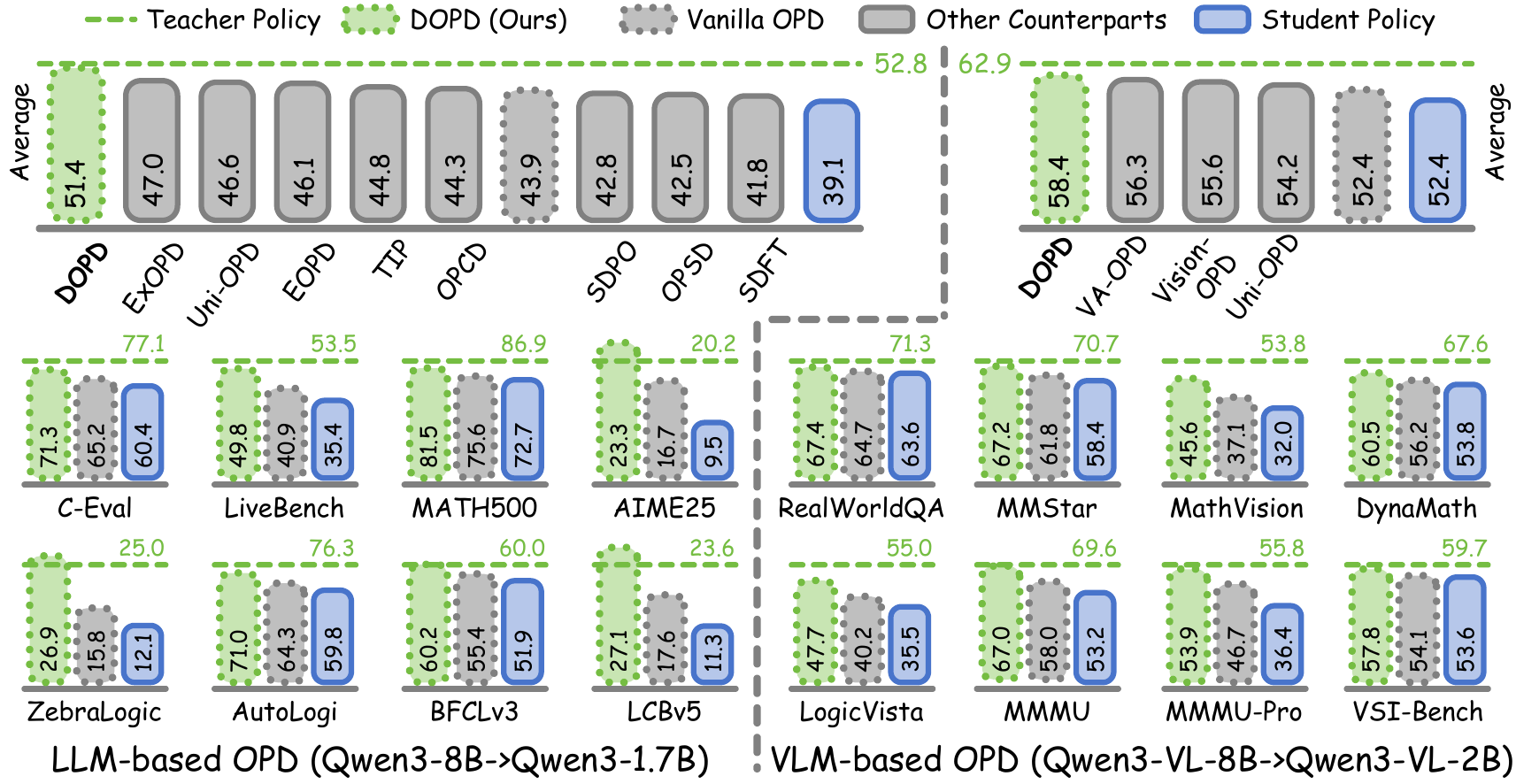}
    \caption{Performance comparison of our DOPD with competing approaches across eight benchmarks in terms of average across all benchmarks (upper bigger bars) and individual values of each benchmark (lower small bars). }
    \label{fig:bar_charts}
\end{figure}

\clearpage

\section{Introduction}
Distillation, as a powerful paradigm for transferring the capabilities of a high-performing teacher policy into a  suboptimal student policy, typically relies on off-policy trajectories, which may expose the student to state distributions that are misaligned with its own evolving behavior~\citep{jin2026entropy,sanh2019distilbert,hsieh2023distilling,yu2026vismem,yao2026joyai}. By contrast, recent OPD paradigms address this limitation by rolling out from the current student policy and using the teacher to provide token-level supervisory signals~\citep{thinking2025opd,song2026survey,xiao2026mimo,xu2026deepseek}. This formulation not only mitigates distribution shift, but also delivers dense per-token teacher supervision via student-sampled on-policy trajectories, yielding higher distillation efficiency and superior performance.

\begin{figure}[t]
    \centering
    \includegraphics[width=0.9\linewidth]{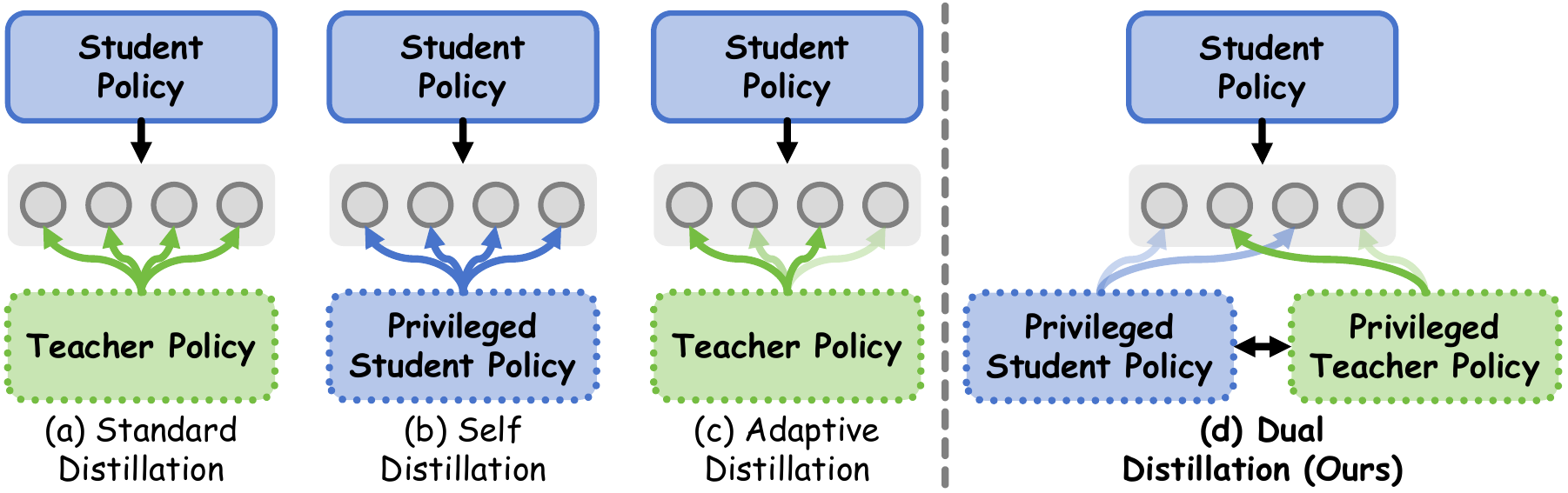}
    \caption{Comparison of existing \textbf{(a)} standard distillation, \textbf{(b)} self distillation, and \textbf{(c)} adaptive distillation paradigms with our proposed \textbf{(d)} dual distillation paradigm.}
    \label{fig:com}
\end{figure}

Although OPD has emerged as an effective post-training paradigm, its achievable upper bound is fundamentally constrained by the quality of the supervision source~\citep{li2026rethinking,fu2026revisiting,ko2026scaling}. As demonstrated in Figure~\ref{fig:com}, for standard strong-to-weak distillation~\citep{ko2025distillm,thinking2025opd,jang2026stable,zhang2026fast,ye2026policy,yang2026learning,hou2026uni,wu2026lightning}, the student is encouraged to imitate a stronger teacher; for self distillation~\citep{shenfeld2026self,hubotter2026reinforcement,zhao2026self,penaloza2026privileged,stein2026gates}, \textit{i.e.}, self-as-teacher pattern, the model improves by regularizing itself under different contexts or conditions. 
In both cases, the effectiveness of OPD implicitly relies on the assumption: \textbf{the supervision signals should reflect a learnable capability beyond the current student policy}. 
However, this assumption might be fragile~\citep{li2026rethinking,song2026survey,yang2026self}, especially when privileged information is introduced.

Privileged information, such as verified reasoning hints for LLMs~\citep{penaloza2026privileged,zhao2026self,hubotter2026reinforcement,yang2026self} or structured visual annotations for VLMs~\citep{yuan2026vision,liu2026visual}, can indeed improve the prediction distribution of teacher policy and raise the apparent ceiling of distillation. 
Nevertheless, the theoretical gains afforded by privileged information training do not necessarily translate into transferable supervisory signals. Rather, they may stem from a hitherto uncharacterized failure mode, \textit{i.e.}, privilege illusion: the ostensible performance gap between teacher and student in fact conflates two fundamentally distinct components. The first is the intrinsic teacher-student capability gap, which is expected to close through distillation; the second is a gap driven by information asymmetry, which arises from the access to privileged inputs that remain almost unlearnable.
Indiscriminately distilling such a teacher distribution may therefore cause the student to fit privileged outcomes rather than acquire transferable ability. 
To summarize, \textbf{adding privileged inputs theoretically improves the ceiling, but the gains may stem from privilege illusion rather than capability optimization}, resulting in rapid entropy collapse, reduced exploration, and ultimately poor distillation effectiveness.

As the distillation signals are highly non-uniform across tokens, the concern of privileged illusion becomes more pronounced.
For realistic trajectories, only a small subset of tokens may encode decisive branches, grounded evidence, critical preferences and other capacity-centric information~\citep{xu2026tip,jin2026entropy}, while many others might provide low-value supervision, which might be privilege-dependent.
However, the vanilla and most variants of OPD methods often optimize all tokens with the same supervision source and objective form, implicitly assuming that each token contributes equally to capability transfer~\citep{thinking2025opd,li2026rethinking,song2026survey}. When incorporating privileged inputs, part of the teacher-student performance advantage originates from information gap rather than transferable capability. In this case, dense supervision might bias the student toward learning privilege-related shortcuts that are easier to fit than the underlying transferable capabilities, thereby amplifying the information-asymmetry component of the teacher-student gap.
Thus, \textbf{indiscriminately and uniformly distilling all tokens from one monolithic policy might intensify the privilege illusion}.

Based on these insights, we propose an advantage-aware dual distillation paradigm, termed as \textbf{DOPD}, exploiting the complementary properties of teacher-based and self-based supervision under the privileged contexts to dynamically route token-level supervision between teacher and student policy according to the \textbf{privilege advantage gap} and their relative predicted probabilities. 
For tokens where the privileged teacher demonstrates a credible capability advantage, we apply stronger teacher distillation to transfer high-value privilege-conditioned capacity. As for tokens that are likely dominated by privileged information or less related to capacity, we instead rely on lighter supervision to preserve stability and encourage favorable exploration. In this way, dual distillation jointly adapts the supervision source, strength, and granularity, enabling more effective, stable, and adaptive distillation with less privilege illusion.

Extensive experiments demonstrate that our method achieves superior distillation performance across a wide range of scenarios, and exhibits excellent robustness, scalability, and generalization.
Specifically, averaged across eight benchmarks, our method outperforms Vanilla OPD by \textbf{7.5} and \textbf{6.0 points} on LLM-based and VLM-based setups respectively, and sustains consistent improvements ranging from \textbf{6.2-10.6 points} across five model pairs of varying sizes.
Furthermore, our method also delivers more favorable performance in continual learning, out-of-distribution evaluation, and training stability. Additional token and divergence analyses, sensitivity and ablation studies further corroborate the effectiveness and rationality of our approach.

\section{Related Works}
\label{sec:ralated_work}

\subsection{Teacher-student Distillability}
Teacher-student distillation has long been studied as a means of transferring capability from a stronger teacher to a weaker student model~\citep{kim2016sequence,sanh2019distilbert,yu2026latent}. In the era of large models, teacher–student distillability has become a more nuanced question than mere teacher imitation: recent work shows that teachers can transfer not only labels, but also rationales, trajectories, preferences, and broader behavioral patterns to smaller students~\citep{hsieh2023distilling,gu2024minillm,agarwal2024policy}. However, such transfer is not monotonic in teacher strength. Studies on the capacity gap suggest that an overly powerful teacher may provide signals that are difficult for a limited student to absorb~\citep{busbridge2025distillation,li2025small,xu2025speculative}. 
Recently, some works also report this phenomenon on OPD settings, indicating that a more compatible initial distribution may be needed for better teacher-student distillation~\citep{li2026rethinking,fu2026revisiting,ko2026scaling,agarwal2024policy}.
Collectively, these studies suggest that effective distillation depends not only on a more powerful teacher model but also on the  content and form of capacity being transferred.

\subsection{On-Policy Distillation}
OPD has emerged as a compelling post-training paradigm that unifies the distributional consistency of on-policy learning with the dense supervision. As depicted in Figure~\ref{fig:com}, this field has evolved along three structured research directions: 
(a) standard distillation, \textit{i.e.}, strong-to-weak paradigm, where a higher-capacity teacher model transfers knowledge to a weaker student via supervision on student-generated rollouts~\citep{agarwal2024policy,ko2025distillm,thinking2025opd}. Recent efforts are primarily  structural modifications to the baseline to enable more stable, faster or more effective: Veto~\citep{jang2026stable}, Fast OPD~\citep{zhang2026fast}, OPCD~\citep{ye2026policy}, ExOPD~\citep{yang2026learning}, Uni-OPD~\citep{hou2026uni}, Lightning OPD~\citep{wu2026lightning}, Vision-OPD~\citep{yuan2026vision}, and VA-OPD~\citep{liu2026visual}. 
(b) self distillation, repurposing a single model as both teacher and student under different context conditions, including: SDFT~\citep{shenfeld2026self}, SDPO~\citep{hubotter2026reinforcement}, OPSD~\citep{zhao2026self}, PI-Distill~\citep{penaloza2026privileged}, RLSD~\citep{yang2026self}, and GATES~\citep{stein2026gates}.
(c) adaptive distillation, which dynamically modulates supervision strategy based on student state, or other training signals: EOPD~\citep{jin2026entropy}, TA-OPD~\citep{wang2026not}, TIP~\citep{xu2026tip}, REOPOLD~\citep{ko2026scaling}, and TSD-KD~\citep{kim2026explain}.

Despite remarkable progress attained by such methods, they remain subject to fundamental limitations.
In Vanilla OPD, student performance is subject to an inherent theoretical ceiling dictated by the performance of teacher policy~\citep{song2026survey,li2026rethinking,thinking2025opd}. This constraint becomes particularly pronounced in challenging tasks, where the teacher itself exhibits subpar performance.
While several lines of research have made preliminary attempts to leverage privileged information~\citep{shenfeld2026self,yang2026self,zhao2026self,hubotter2026reinforcement,penaloza2026privileged,stein2026gates,yuan2026vision,liu2026visual}, these approaches generally operate under the implicit assumption that transferable capabilities can be enhanced via the direct integration of privileged information, and the supervision signals should be received uniform distillation mechanisms consistently from monolithic source. Critically, these methods fundamentally overlook the risk of privilege illusion, thus may fail to explicitly identify and distill genuine inherent capacity.

\section{Methodology}

\begin{figure}[t]
    \begin{subfigure}{0.48\linewidth}
        \centering
        \includegraphics[width=\linewidth]{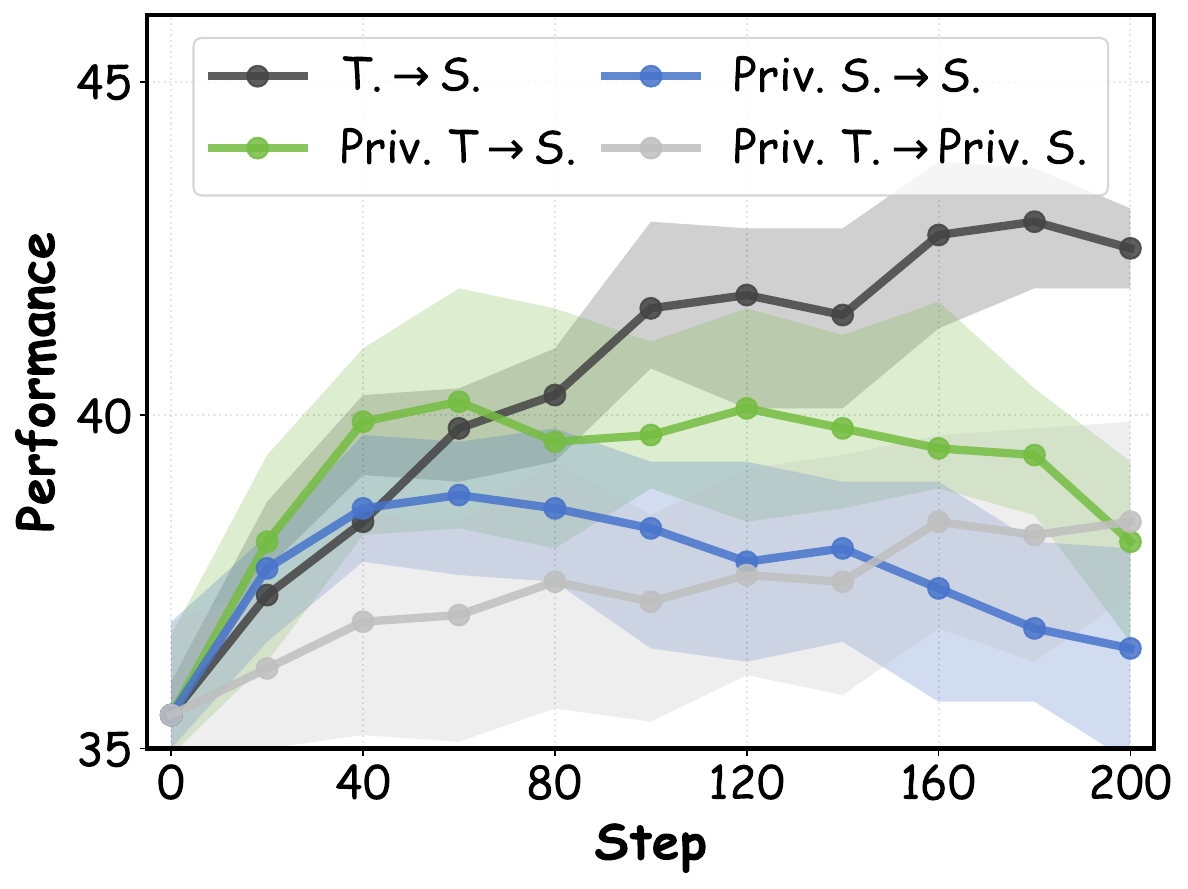}
        \caption{Performance vs. Training Step}
        \label{fig:illusion1}
    \end{subfigure}
    \hfill
    \begin{subfigure}{0.48\linewidth}
        \centering
        \includegraphics[width=\linewidth]{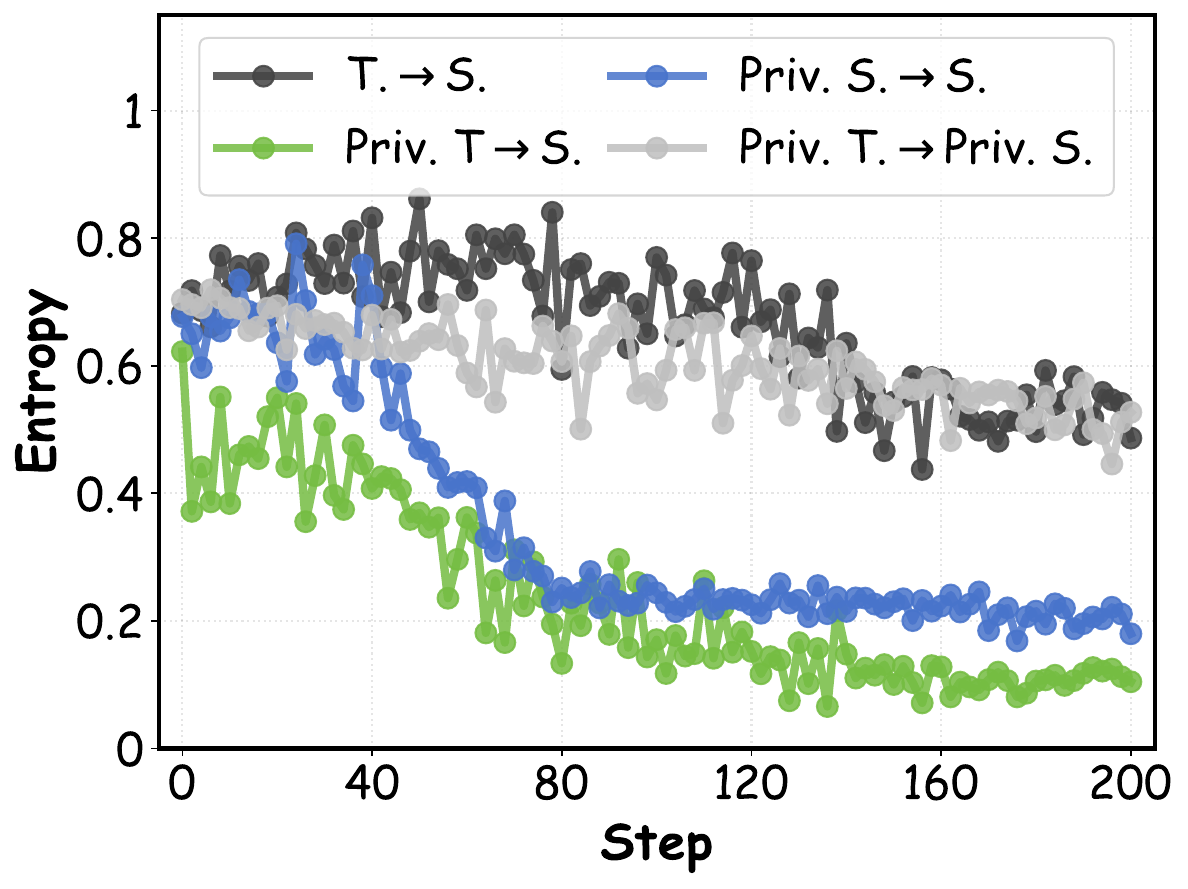} 
        \caption{Entropy vs. Training Step}
        \label{fig:illusion2}
    \end{subfigure}
    
    \caption{Comparison of \textbf{(a)} performance and \textbf{(b)} entropy on OPD variants with privileged information. Here, T., S., and Priv. denote teacher policy, student policy and with privileged information, respectively.}
    \label{fig:illusion}
\end{figure}

\subsection{Background}
\label{sec:background}
\subsubsection{Privilege Illusion}
Existing OPD fundamentally relies on the assumption that a stronger teacher provides richer and more informative supervision~\citep{thinking2025opd,ko2025distillm}. 
Thus, in many practical scenarios, an intuitive exploration is to equip teachers or student itself with privileged inputs~\citep{song2026survey}. 
For instance, verified hints in reasoning-centric tasks~\citep{zhao2026self,penaloza2026privileged,hubotter2026reinforcement}, or bounding boxes of objects in visual perception scenarios~\citep{liu2026visual,yuan2026vision}.
Here, as exampled in Figure~\ref{fig:privilege_demo_llm} and \ref{fig:privilege_demo_vlm}, we employ a moderate form of privileged information that delivers essential cues while refraining from directly disclosing detailed execution procedures and final answers (influence of various forms of privileged information will be discussed in Section~\ref{sec:pi_analysis}).
However, when augmented with privileged information, the prediction advantage may arise from information asymmetry rather than genuine inherent capability. Uncurated distilling such signals can encourage the student to imitate privileged outcomes instead of acquiring practical and transferable abilities, or triggers distillability due to irreparable teacher-student gap, leading to inferior and unstable distillation process, and unfavorable entropy collapse~\citep{cui2025entropy,yu2026dapo}.

As illustrated in Figure~\ref{fig:illusion}, we compare the impact of privileged information inclusion on both performance  and entropy trends. We evaluate three OPD variants, in which privileged information is granted to the teacher policy only, the student policy only, and both policies, respectively. We observe that introducing privileged information to either the teacher or the student separately delivers modest performance improvements over Vanilla OPD in the very early training phase, yet the information asymmetry between the two policies gives rise to late-stage performance degradation coupled with entropy collapse. 
When both policies are granted access to privileged information, the superficial advantage conferred by information asymmetry vanishes. Furthermore, uniform distillation across all tokens under this setting fails to enable the student to genuinely internalize the core competencies. Instead, the student merely passively adapts to  the privileged information, ultimately yielding only marginal performance improvements less than Vanilla OPD.

In summary, the results reveal that straightforward incorporation of privileged information might create a failure phenomenon termed privilege illusion: privileged inputs may yield an ostensible advantage, however, such gains often stem from information asymmetry rather than from a genuine enhancement of capability.

\begin{figure}[t]
    \begin{subfigure}{0.48\linewidth}
        \centering
        \includegraphics[width=\linewidth]{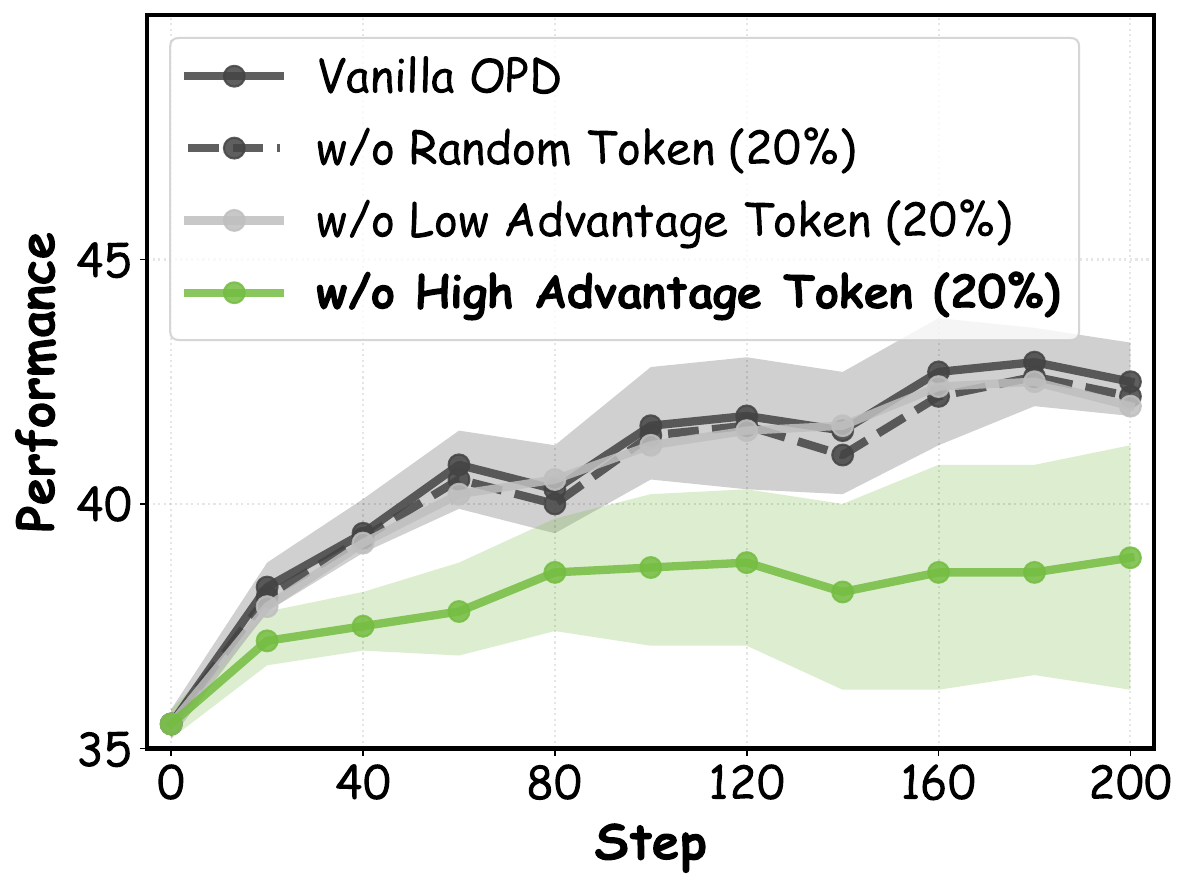}
        \caption{Qwen3-8B $\to$ Qwen3-1.7B (LiveBench)}
        \label{fig:background1}
    \end{subfigure}
    \hfill
    \begin{subfigure}{0.48\linewidth}
        \centering
        \includegraphics[width=\linewidth]{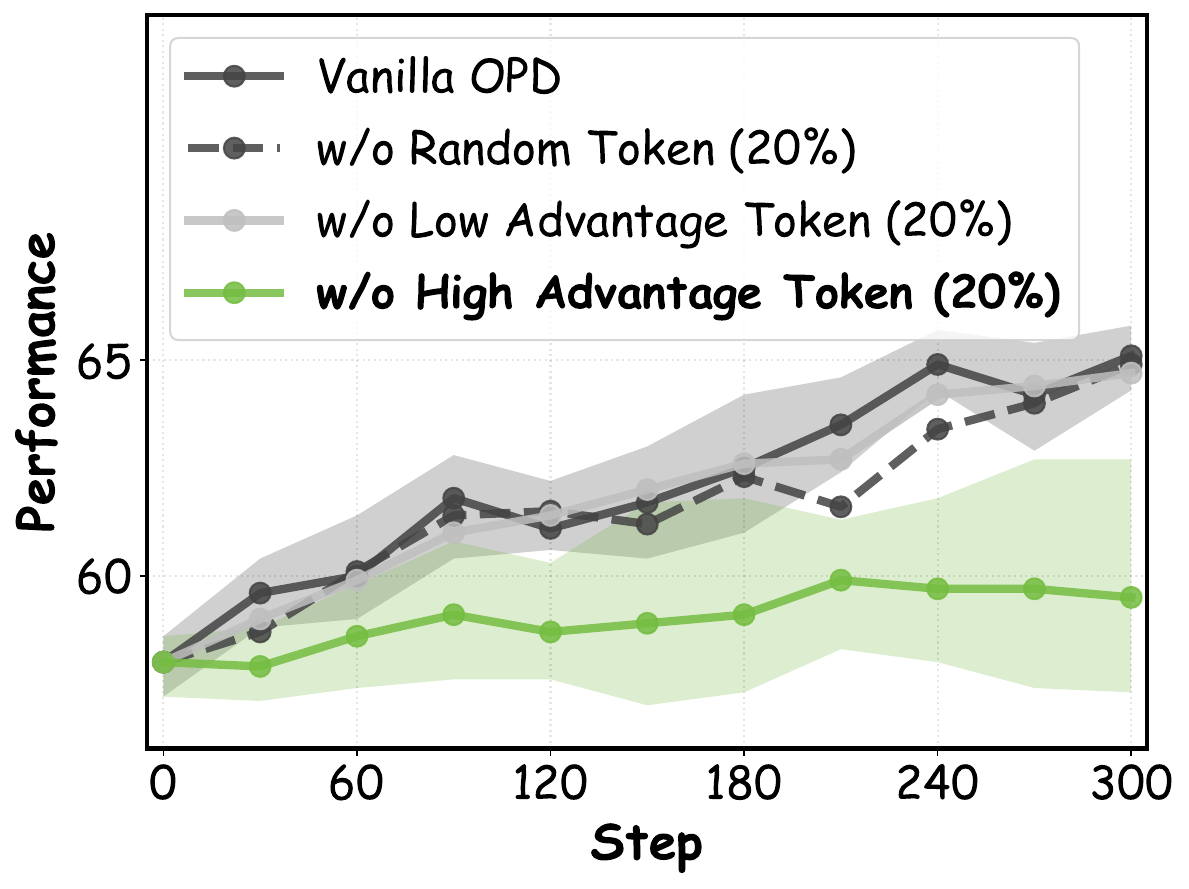} 
        \caption{Qwen3-VL-8B $\to$ Qwen3-VL-2B (MMStar)}
        \label{fig:background2}
    \end{subfigure}
    
    \caption{Token ablations on random tokens, and tokens with high or low advantage gap.}
    \label{fig:background}
\end{figure}

\subsubsection{Privilege Advantage Gap}
\label{sec:advantage_gap}

As mentioned above, a key limitation of existing OPD methods is their inability to disentangle capability gaps from information gaps. 
Thus, we argue that, when both with privileged inputs, the relative advantage between a teacher policy and a student policy offers a proxy for privilege-conditioned prediction gap. Consequently, a large advantage gap indicates capability discrepancy under controlled privileged conditions, whereas a small gap suggests that the advantage of teacher policy is primarily attributable to privileged information. This perspective motivates a privilege advantage-aware distillation paradigm that selectively transfers knowledge when the supervision signal reflects authentic competence rather than privilege illusion.

For a given original input $\mathbf{x}$, the student policy samples an output sequence from the conditional distribution. 
To conduct privilege advantage-aware distillation, we aim to investigate the distribution disparity between the teacher policy $\Pi_{T}$ and student policy $\Pi_{S}$ when both have access to privileged inputs, termed the privilege advantage gap $\mathcal{A}$. Then, we perform forward passes on the two policies respectively, and take the absolute value of their log-probability difference as the final privilege advantage gap:
\begin{equation}
    \mathcal{A} = | \log \Pi_{T} \left(\mathbf{y}_{n} \mid \mathbf{x} , \mathbf{p} , \mathbf{y}_{<n}\right)-\log \Pi_{S}\left(\mathbf{y}_{n} \mid \mathbf{x}, \mathbf{p}, \mathbf{y}_{<n}\right)| = \left|\log\frac{  \Pi_{T} \left(\mathbf{y}_{n} \mid \mathbf{x} , \mathbf{p} , \mathbf{y}_{<n}\right)}{ \Pi_{S}\left(\mathbf{y}_{n} \mid \mathbf{x}, \mathbf{p}, \mathbf{y}_{<n}\right)}\right|,
    \label{eq:advantage_gap}
\end{equation}
where $\mathbf{y}_{n}$ denotes the current token to be evaluated by the two policies, and $\mathbf{p}$ denotes the privileged information provided as auxiliary contexts along with previous tokens. 
The quantity $\mathcal{A}$ captures the prediction discrepancy stemming from the performance gap between teacher and student policies under identical privileged conditions, which constitutes the idealized learning content.

To further verify the rationality of privilege advantage gap to separate capacity and information gap, we conduct comprehensive ablation studies and empirical analyses across both large LLMs and VLMs. Specifically, we construct three variants of the Vanilla OPD paradigm, each discarding particular tokens without distillation loss: (1) a reference baseline that randomly drops 20\% of tokens; (2) a variant that prunes the 20\% of tokens with the smallest advantage gap; (3) a variant that prunes the 20\% of tokens with the largest advantage gap.
As illustrated in Figure~\ref{fig:background}, ablating high-advantage tokens incurs substantial performance degradation and a marked reduction in distillation efficiency. At the 100th optimization step, removing high-advantage tokens achieves only approximately 50\% of the performance gain obtained by Vanilla OPD. In contrast, pruning random or low-advantage tokens exerts negligible performance impact relative to the Vanilla OPD baseline. This performance disparity is even more pronounced in multimodal models, achieves only about 20\% of the improvement achieved by Vanilla OPD. It is also noteworthy that despite underperforming relative to all counterparts, the variant with high-advantage tokens removed still yields tangible distillation gains by 3.4 and 1.5 points, indicating that the remaining tokens, though less critical, remain indispensable to distillation.

\subsubsection{Takeaway}
Based on these backgrounds, we summarize the following takeaway to support our proposed method:
\begin{takeaway}
Naively injecting privileged information can create a privilege illusion, where apparent gains arise from information asymmetry instead of transferable capability. The privilege advantage gap could highlight high-value tokens whose supervision is most critical for capacity-centric distillation.
\end{takeaway}

\subsection{DOPD: Dual On-policy Distillation}
\label{sec:dopd}
\subsubsection{Divergence}
As discussed in prior work~\citep{hubotter2026reinforcement,jin2026entropy,li2026rethinking}, to learn a student from a teacher under the OPD framework, we first consider three common divergence-based objectives derived from Kullback-Leibler (KL) divergence: forward KL, reverse KL, and Jensen-Shannon (JS) divergence. For notational simplicity, we omit the distinction between privileged and non-privileged observations and denote $\mathbf{t}$ as the current contexts of the teacher and student policies.

\noindent\textbf{Forward KL Divergence.}
It encourages the student to cover the full support of the teacher distribution by penalizing actions that receive non-negligible probability under the teacher but are underestimated by the student. As a result, this objective promotes comprehensive imitation of the action preferences of teacher:
\begin{equation}
    KL_{\mathrm{forward}}\left(\Pi_{T}\,\middle\|\,\Pi_{S}\right) = \mathbb{E}_{\mathbf{y} \sim \Pi_{T}\left(\cdot \mid \mathbf{t}\right)} \left[ \log \frac{ \Pi_{T}\left(\mathbf{y} \mid \mathbf{t}\right) }{ \Pi_{S}\left(\mathbf{y} \mid \mathbf{t}\right) } \right].
\end{equation}

\noindent\textbf{Reverse KL Divergence.}
It encourages the student to concentrate probability mass on actions strongly favored by the teacher, while assigning little emphasis to low-probability regions of the teacher distribution. Such mode-seeking behavior often leads to sharper student policies, but may also discard informative secondary modes encoded by the teacher:
\begin{equation}
    KL_{\mathrm{reverse}} \left(\Pi_{S}\,\middle\|\,\Pi_{T}\right) = \mathbb{E}_{\mathbf{y} \sim \Pi_{S}\left(\cdot \mid \mathbf{t}\right)} \left[ \log \frac{ \Pi_{S}\left(\mathbf{y} \mid \mathbf{t}\right) }{ \Pi_{T}\left(\mathbf{y} \mid \mathbf{t}\right) } \right].
\end{equation}

\noindent\textbf{JS Divergence.}
It introduces an intermediate average distribution, and calculate the KL divergence of teacher and student relative to this medium, without directional bias in forward or reverse directions. 
Their combination provides a more balanced optimization signal, thereby improving the stability of policy distillation:
\begin{equation}
    JS = \frac{1}{2} KL\left( \Pi_T \,\middle\|\, \Pi_M \right) + \frac{1}{2}KL\left(\Pi_S\,\middle\|\,\Pi_M\right),
    \text{where} \quad \Pi_M = \frac{1}{2} \Pi_T + \frac{1}{2} \Pi_S.
\end{equation}

\subsubsection{OPD}
As a promising post-training paradigm, OPD holds its core advantage in performing knowledge transfer with samples drawn from the target student policy to effectively mitigate performance bias caused by distribution shift, and provide richer supervision signals than conventional reinforcement learning paradigms~\citep{thinking2025opd,song2026survey,li2026rethinking}. Specifically, given the student policy, for particular inputs $\mathbf{x}$, it samples the sequence of predicted trajectory $\mathbf{y}\sim\Pi_{S}\left(\cdot\mid\mathbf{x}\right)$. Then, the teacher policy, typically a stronger model, will offer token-level signals as optimization. Thus, the optimization objective of Vanilla OPD could be summarized as:
\begin{equation}
    \mathbb{E}_{\mathbf{x} \sim \mathcal{D}}\left[\mathbb{E}_{\mathbf{y} \sim \Pi_{S}}\left[\frac{1}{|\mathbf{y}|} \sum_{n=1}^{|\mathbf{y}|} \mathcal{L}_{n}\left(\mathbf{y}_{n} ; \mathbf{t}_{<n}\right)\right]\right],
    \label{eq:objective}
\end{equation}
where $\mathbf{t}$ denotes the conditioning context, which comprises the original inputs, previously generated tokens, and auxiliary information if available, and $\mathcal{L}$ quantifies the token-level divergence between the teacher and student policies. Conventionally, this penalty term takes the form of divergence-based objectives, \textit{e.g.}, widely adopted reverse KL, as well as alternative divergence variants or combinations thereof. Fundamentally, nearly all advancements in OPD center on minimizing the objective formalized in Equation~\ref{eq:objective}, so as to yield a student model whose behavioral distribution aligns more closely with that of the teacher.

In addition, OPD approaches have different granularity of teacher supervision, ranging from coarse to fine: sampled-token, Top-K token, and full-vocabulary distillation.
Sampled-token distillation confines its distillation objective exclusively to the predicted target token, while Top-K token distillation expands the scope of supervision to cover the k tokens with the highest predictive probabilities.
By contrast, full-vocabulary distillation aligns the complete probability distribution across the entire vocabulary.
The density of informative supervisory signals increases monotonically, which theoretically leads to higher efficiency,  however, this gain comes at the cost of higher computational overhead and potential risk of training instability, which stems from overfitting to the inherently noisy distributions of low-probability tokens~\citep{zhao2026self,li2026rethinking,hubotter2026reinforcement}.
Accordingly, the selection of distillation paradigm in practical deployment is typically tailored to specific downstream objectives and computational budgets.

\begin{figure}[t]
    \centering
    \includegraphics[width=0.9\linewidth]{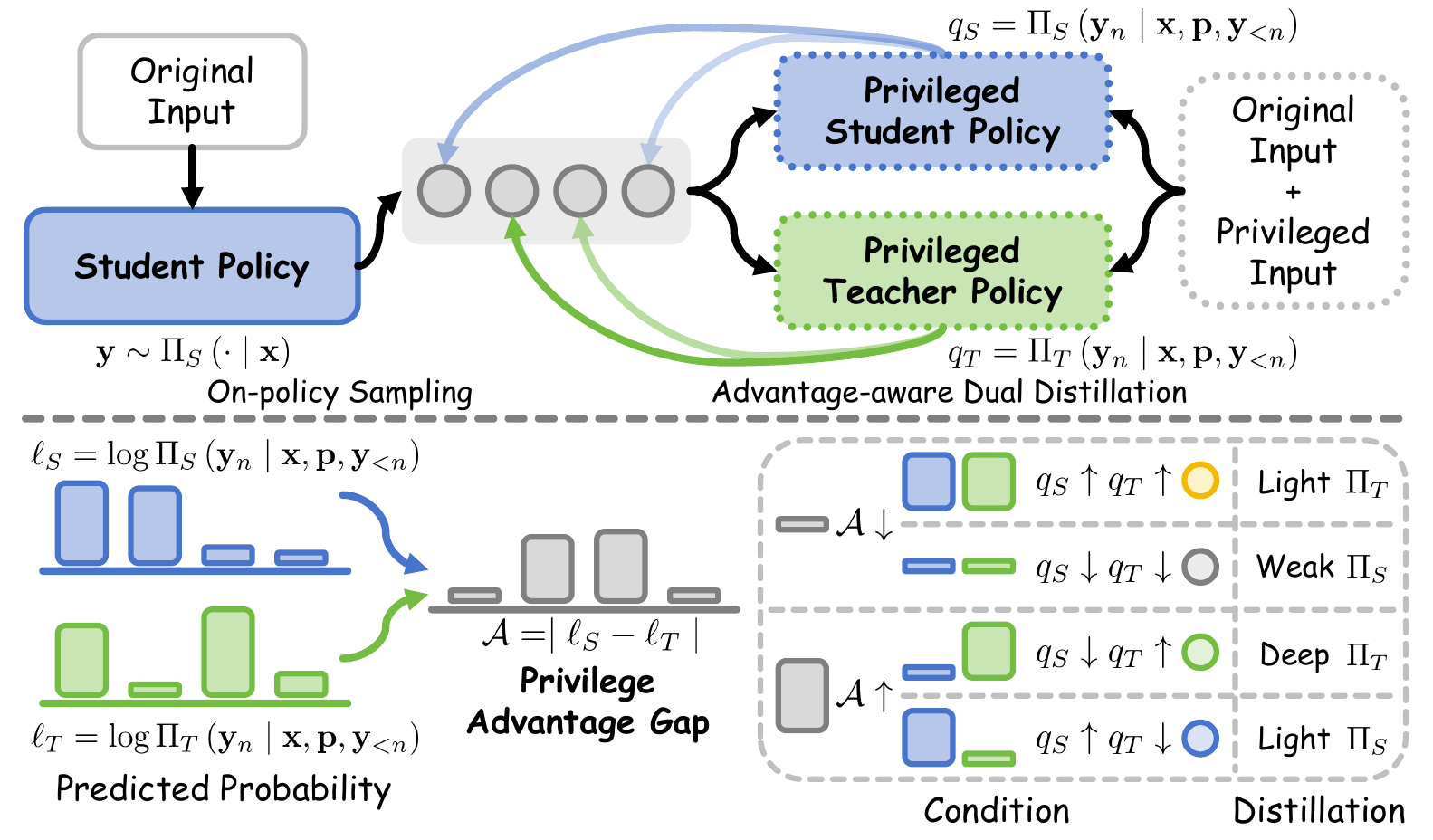}
    \caption{Overview of our proposed DOPD.}
    \label{fig:overview}
\end{figure}

\subsubsection{Advantage-aware Dual Distillation}

As discussed above in Section~\ref{sec:background}, not all tokens should receive supervision of identical objective and strength or from the same source. 
When privileged information is introduced, the apparent superiority of teacher may originate either from privilege-conditioned capability discrepancy or from information asymmetry. 
Therefore, indiscriminately distilling the privileged teacher distribution may transfer shortcut-like privileged cues, while overly conservative self-teaching may fail to capture genuinely beneficial knowledge. 
To address this issue, we propose advantage-aware dual distillation, which dynamically selects both the supervision source and the distillation form according to the token-level privilege advantage gap.

Concretely, for each on-policy sampling trajectory $\mathbf{y}$, we perform additional privileged forward passes: one with the privileged student policy and the other with the privileged teacher policy. For the n-\textit{th} token, we denote their token-level probabilities as: $q_{S}= \Pi_{S} \left( \mathbf{y}_{n} \mid \mathbf{x} , \mathbf{p}, \mathbf{y}_{<n} \right)$ and $ q_{T}= \Pi_{T} \left( \mathbf{y}_{n} \mid \mathbf{x} , \mathbf{p}, \mathbf{y}_{<n} \right)$, while corresponding token-level log-probabilities as:  $\ell_{S}= \log \Pi_{S} \left( \mathbf{y}_{n} \mid \mathbf{x} , \mathbf{p}, \mathbf{y}_{<n} \right)$ and $ \ell_{T}= \log \Pi_{T} \left( \mathbf{y}_{n} \mid \mathbf{x} , \mathbf{p}, \mathbf{y}_{<n} \right)$. 
Here, the privileged student policy shares parameters with the deployed student policy, but receives the privileged input $\mathbf{p}$ during training, while the privileged teacher policy remains frozen. As we formally defined in Equation~\ref{eq:advantage_gap} of Section~\ref{sec:advantage_gap}, we use the two privileged policies to calculate the privilege advantage gap $\mathcal{A}$. For a scored n-\textit{th} token, we compare its $\mathcal{A}_{n}$, $q_{S}$, and $q_{T}$  with their average $\bar{\mathcal{A}}$, $\bar{q_{S}}$, and $\bar{q_{T}}$, respectively. In practice, to ensure stability, we first discard the top 5\% of outliers and perform normalization within the batch, then use them to calculate the average. 
Based on all these relationships, it yields four token regimes, each corresponding to a distinct learning strategy.

\noindent\textbf{Low $\mathcal{A}$ with High $q_{S}$ \& $q_{T}$.}
When the two privileged policies have low advantage gap with both high predicted probability, \textit{i.e.}, $\mathbb{I}^{\mathrm{LH}}=\left(\mathcal{A}_{n}< \bar{\mathcal{A}}\right) \wedge \left(q_{S} + q_{T} \geq \bar{q_{S}} + \bar{q_{T}}\right)$, the privileged teacher and privileged student make consistent and confident predictions. 
In this case, the bottleneck is mainly attributed to the absence of privileged information rather than an inherent capability gap. 
Thus, directly enforcing full teacher imitation is unnecessary and may over-transfer privileged shortcuts. 
We instead apply a light teacher distillation objective, using Top-K reverse KL to absorb useful privileged knowledge in a conservative manner:
\begin{equation}
    \mathcal{L}^{LH}
    =
    \beta_{l}\,
    KL_{\mathrm{reverse}}
    \left(
        \Pi_{S}\left(\cdot\mid \mathbf{x},\mathbf{y}_{<n}\right)
        \,\middle\|\,
        \Pi_{T}\left(\cdot\mid \mathbf{x},\mathbf{p},\mathbf{y}_{<n}\right)
    \right).
\end{equation}
where $\beta_{l}$ denotes the intensity coefficient with light distillation.

\noindent\textbf{Low $\mathcal{A}$ with Low $q_{S}$ \& $q_{T}$.}
When the two privileged policies have low advantage gap with both low predicted probability, \textit{i.e.}, $\mathbb{I}^{\mathrm{LL}}=\left(\mathcal{A}_{n}< \bar{\mathcal{A}} \right)\wedge \left(q_{S} + q_{T} < \bar{q_{S}}  + \bar{q_{T}}\right)$, both privileged policies assign low probability to the current token. 
Such tokens are likely to lie beyond the reliable competence region of both models, where aggressive teacher forcing may introduce noisy or even misleading supervision. 
Therefore, we use the privileged student as a weak self-regularizing anchor, using Top-K reverse KL with a smaller coefficient, to stabilize training without forcing the student to imitate uncertain teacher predictions:
\begin{equation}
    \mathcal{L}^{LL}
    =
    \beta_{w}\,
    KL_{\mathrm{reverse}}
    \left(
        \Pi_{S}\left(\cdot\mid \mathbf{x},\mathbf{y}_{<n}\right)
        \,\middle\|\,
        sg\left[
        \Pi_{S}\left(\cdot\mid \mathbf{x},\mathbf{p},\mathbf{y}_{<n}\right)
        \right]
    \right),
\end{equation}
where $sg[\cdot]$ denotes stop-gradient to avoid changing the gradient simultaneously to cause target drift, $\beta_{w}$ denotes the intensity coefficient with weak distillation, and $\beta_{w} < \beta_{l}$. In this regime, the privileged student is not treated as a knowledge source, but as a parameter-shared consistency anchor that prevents policy drift.

\noindent\textbf{High $\mathcal{A}$ with High $q_{T}$.}
When the two privileged policies have high advantage gap with high predicted probability of the teacher policy, \textit{i.e.}, $\mathbb{I}^{\mathrm{HT}}=\left(\mathcal{A}_{n} \geq \bar{\mathcal{A}} \right) \wedge \left(q_{T} \geq q_{S}\right)$, the privileged teacher exhibits a clear and confident advantage over the privileged student. 
Since both policies observe the same privileged information, a large privilege advantage gap suggests that, the teacher provides a potentially useful capability signal beyond what the student currently captures.
Accordingly, these tokens contain critical transferable knowledge and should receive stronger supervision. 
We therefore perform full-vocabulary teacher distillation with unit weight, using JS divergence to balance support coverage and mode concentration:
\begin{equation}
    \mathcal{L}^{HT}
    =
    JS
    \left(
        \Pi_{S}\left(\cdot\mid \mathbf{x},\mathbf{y}_{<n}\right)
        \,\middle\|\,
        \Pi_{T}\left(\cdot\mid \mathbf{x},\mathbf{p},\mathbf{y}_{<n}\right)
    \right).
\end{equation}

Compared with Top-K strategy, full-vocabulary alignment provides denser distributional signals, enabling the student to acquire both dominant decisions and informative secondary preferences from the teacher.

\noindent\textbf{High $\mathcal{A}$ with High $q_{S}$.}
When the two privileged policies have high advantage gap with high predicted probability of the student policy, \textit{i.e.}, $\mathbb{I}^{\mathrm{HS}}=\left(\mathcal{A}_{n} \geq \bar{\mathcal{A}} \right)\wedge \left(q_{T} < q_{S}\right)$, the privileged student assigns relative larger confidence while the privileged teacher does not provide a comparably reliable signal. 
In this regime, strongly constraining the student toward the teacher may suppress potentially valid exploratory behavior. 
We therefore adopt a light privileged-student distillation objective with Top-K reverse KL, which softly encourages consistency between the deployed student and its privileged counterpart while avoiding over-regularization:
\begin{equation}
    \mathcal{L}^{HS}
    =
    \beta_{l}\,
    KL_{\mathrm{reverse}}
    \left(
        \Pi_{S}\left(\cdot\mid \mathbf{x},\mathbf{y}_{<n}\right)
        \,\middle\|\,
        sg\left[
        \Pi_{S}\left(\cdot\mid \mathbf{x},\mathbf{p},\mathbf{y}_{<n}\right)
        \right]
    \right).
\end{equation}

\noindent\textbf{Total Objective.}
Finally, we combine the four token-wise objectives through indicator masks:
\begin{equation}
\begin{aligned}
    \mathcal{L}^{\mathrm{DOPD}}
    =&\;
    \mathbb{I}^{\mathrm{LH}}\mathcal{L}^{\mathrm{LH}}
    +
    \mathbb{I}^{\mathrm{LL}}\mathcal{L}^{\mathrm{LL}}
    +
    \mathbb{I}^{\mathrm{HT}}\mathcal{L}^{\mathrm{HT}}
    +
    \mathbb{I}^{\mathrm{HS}}\mathcal{L}^{\mathrm{HS}},
\end{aligned}
\end{equation}
where the masks are determined by the privilege advantage gap and relative probability comparisons described above, which exhaustively partitions the token space under the defined conditions. Thus, the overall optimization objective could be formulated as:
\begin{equation}
    \mathbb{E}_{\mathbf{x} \sim \mathcal{D}}\left[\mathbb{E}_{\mathbf{y} \sim \Pi_{S}}\left[\frac{1}{|\mathbf{y}|} \sum_{n=1}^{|\mathbf{y}|} \mathcal{L}_{n}^{DOPD}\left(\mathbf{y}_{n} ; \mathbf{x}, \mathbf{p}, \mathbf{y}_{<n}\right)\right]\right],
\end{equation}

Through this adaptive routing mechanism, DOPD assigns strong full-vocabulary teacher supervision only to tokens where the privileged teacher demonstrates a credible capability advantage, applies light teacher distillation when the signal mainly reflects privileged information, relies on weak privileged-student regularization for uncertain regions, and preserves student exploration when the privileged student is already confident. 
Consequently, the proposed objective mitigates the entanglement between capability transfer and privileged-information imitation, yielding a more selective, stable, and generalizable OPD paradigm.

\section{Experiments}
\subsection{Settings}
\subsubsection{Models}
We perform all the experiments on Qwen3~\citep{yang2025qwen3} and Qwen3-VL~\citep{bai2025qwen3} families of non-thinking versions as both teacher and student policies. Specifically, the main experiments and all analyses are conducted on Qwen3-8B to Qwen3-1.7B pair, and for VLM scenario is based on Qwen3-VL-8B to Qwen3-VL-2B pair. Besides, to verify the generalization ability of our method, we also add Qwen3-8B to Qwen3-0.6B, Qwen3-4B to Qwen3-0.6B, Qwen3-4B to Qwen3-1.7B, and Qwen3-1.7B to Qwen3-0.6B pairs.

\begin{figure}[h]
    \centering
    \includegraphics[width=0.95\linewidth]{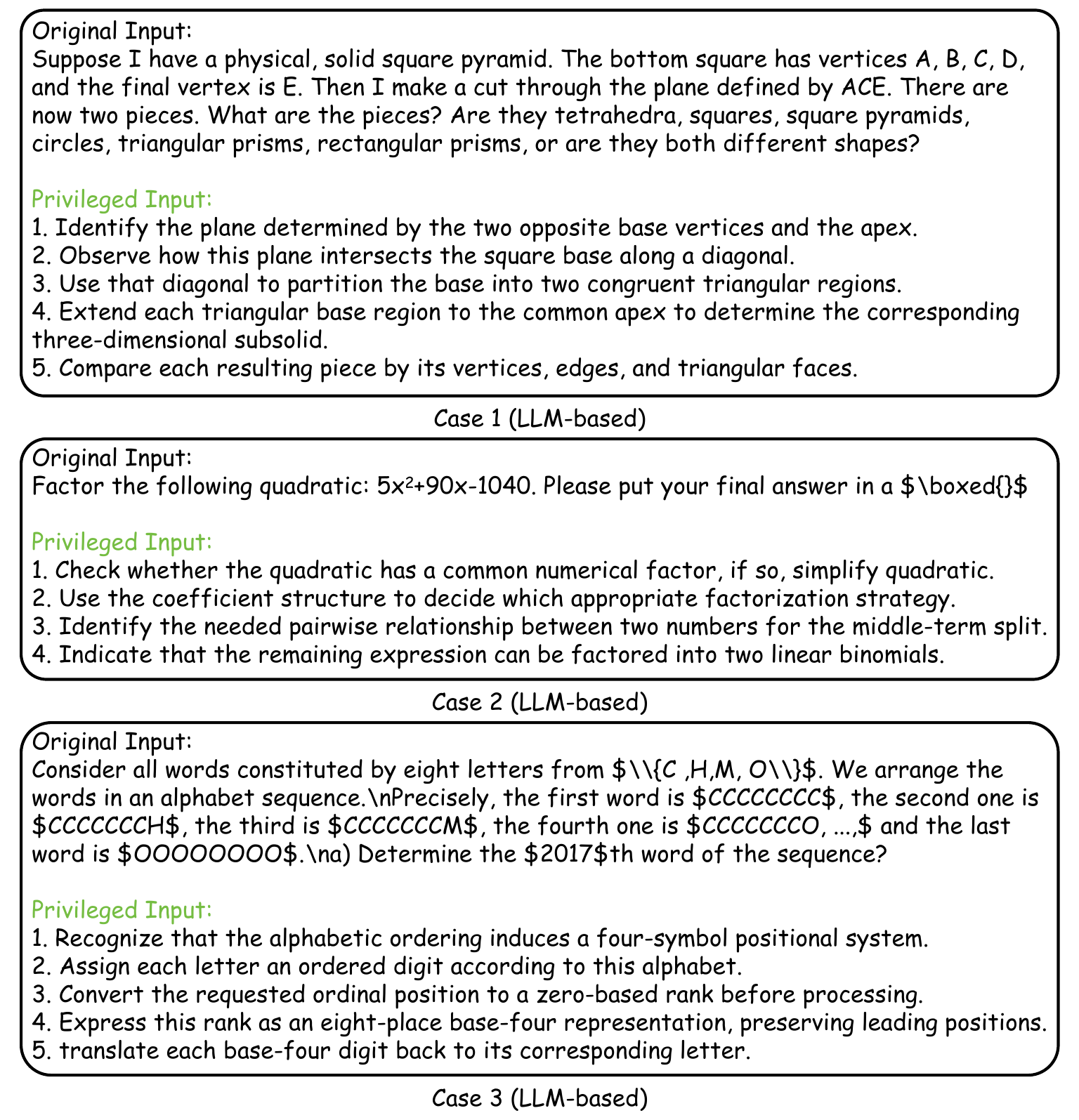}
    \caption{Demonstrations of LLM-based privileged input.}
    \label{fig:privilege_demo_llm}
\end{figure}

\begin{figure}[h]
    \centering
    \includegraphics[width=0.95\linewidth]{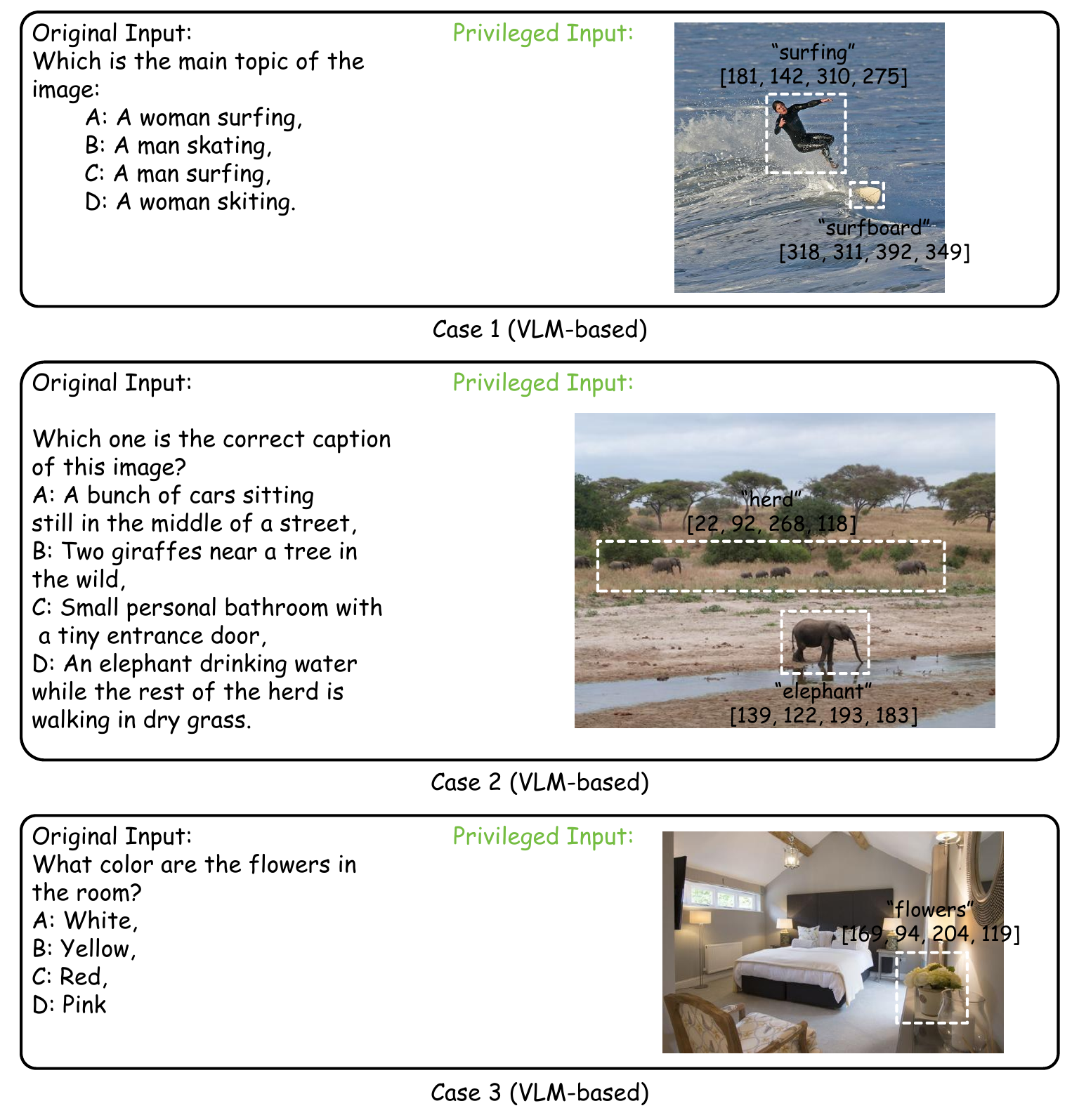}
    \caption{Demonstrations of VLM-based privileged input.}
    \label{fig:privilege_demo_vlm}
\end{figure}

\begin{figure}[h]
    \centering
    \includegraphics[width=0.95\linewidth]{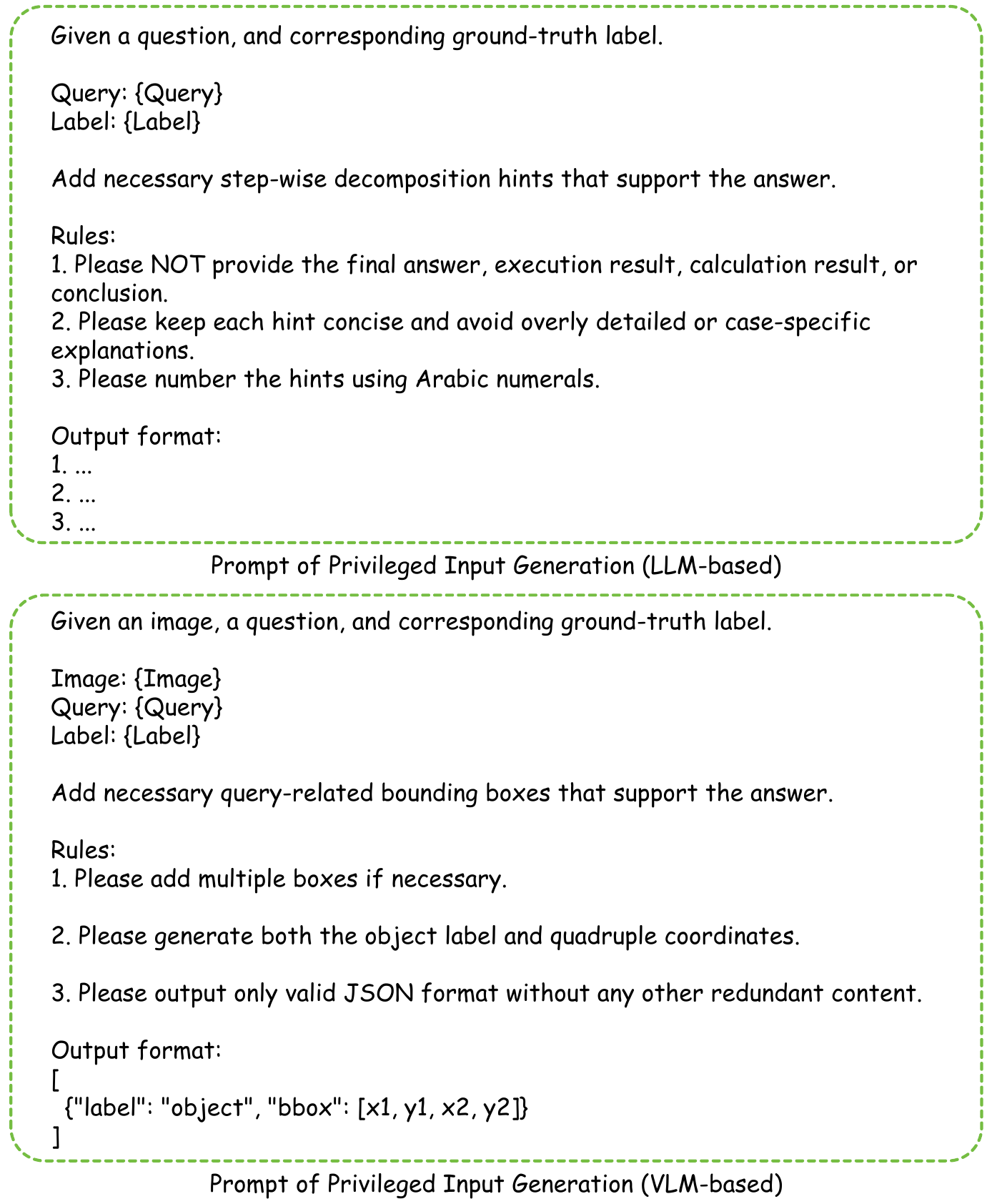}
    \caption{Prompts of Privileged Input Generation.}
    \label{fig:privilege_prompt}
\end{figure}

For the training datasets of LLM-based OPD, we use the high-quality mixture dataset from RaR-Science-20K~\citep{gunjal2025rubrics}, DAPO-Math-17K~\citep{yu2026dapo}, and Skywork-OR1-Coding-14K~\citep{he2025skywork}, covering general, reasoning, and coding tasks. For VLM-based training datasets, we utilize ViRL39K~\citep{li2026lavida} dataset, covering general, visual reasoning and visual understanding tasks. For the corresponding privileged input, we use GPT-5.4~\citep{gpt54} (2026-03-05) to generate step-wise decomposition hints and structured visual annotations respectively, where the generation prompts are provided in Figure~\ref{fig:privilege_prompt}.
As illustrated in Figure~\ref{fig:privilege_demo_llm}, for LLM tasks, we utilize verified rationales as privileged information, with step-wise decomposition hints, but without direct execution trace or final answer.
While as shown in Figure~\ref{fig:privilege_demo_vlm}, for VLM tasks, privileged information denotes structured visual annotations, here we use query-related bounding boxes, with object labels and quadruple coordinates to provide explicit visual context.
To guarantee the data quality, we use GPT-5.4 again to recheck the generated privileged contents, and directly discard relatively low-quality samples, eventually resulting in 32K and 25K high-quality training data for LLM and VLM, respectively.

\subsubsection{Benchmarks}
To evaluate the effectiveness of our method, we employ eight benchmarks for LLM-based OPD, covering three core abilities: (1) general: C-Eval~\citep{huang2023c}, and LiveBench~\citep{white2024livebench}; (2) reasoning: MATH500~\citep{hendrycks2021measuring}, AIME25~\citep{aime25aimi}, ZebraLogic~\citep{lin2025zebralogic}, and AutoLogi~\citep{zhu2025autologi}; and (3) coding: BFCLv3~\citep{bfcl24bfcl}, and LCBv5~\citep{jain2025livecodebench}. For VLM-based OPD, we also include eight benchmarks on three aspects: (1) general: RealWorldQA~\citep{realworld24real}, and MMStar~\citep{chen2024we}; (2) visual reasoning: MathVision~\citep{wang2024measuring}, DynaMath~\citep{zou2025dynamath}, and LogicVista~\citep{xiao2024logicvista}; and (3) visual understanding: MMMU~\citep{yue2024mmmu}, MMMU-Pro~\citep{yue2025mmmu}, and VSI-Bench~\citep{yang2025thinking}. All benchmarks are evaluated using their official metrics and evaluations to ensure fair and consistent comparison.

\begin{table*}[t]
\centering
\caption{Performance comparison of our proposed DOPD with counterparts on general, reasoning, and coding tasks. $\Delta$ indicates the performance gap between the student and teacher policies with \textcolor{gray!60}{gray} cells, while \textcolor{myblue!80}{blue} for over 50\% mitigation and \textcolor{mygreen!80}{green} for complete gap removal by employing the OPD paradigms. The best and second best values are \textbf{bolded} and \underline{underlined}, respectively.}
\setlength{\tabcolsep}{0.9mm}
\resizebox{1\textwidth}{!}{
\begin{tabular}{l|c|cccccccc|c}
\toprule   
\multirow{2}{*}{\textbf{Method}} & \multirow{2}{*}{\textbf{Type}} &\multicolumn{2}{c}{\textbf{General}} & \multicolumn{4}{c}{\textbf{Reasoning}} & \multicolumn{2}{c|}{\textbf{Coding}} & \multirow{2}{*}{\textbf{Average}} \\ 
\cmidrule(lr){3-4} \cmidrule(lr){5-8} \cmidrule(lr){9-10}     
& & \textbf{C-Eval} & \textbf{LiveBench} &  \textbf{MATH500} & \textbf{AIME25} & \textbf{ZebraLogic} & \textbf{AutoLogi} & \textbf{BFCLv3}  & \textbf{LCBv5} &  \\  \midrule
Teacher Policy & \multirow{3}{*}{-} & 77.1 & 53.5 & 86.9 & 20.2 & 25.0 & 76.3 & 60.0 & 23.6 & 52.8 \\
Student Policy &  & 60.4 & 35.4 & 72.7 & 9.5 & 12.1 & 59.8 & 51.9 & 11.3 & 39.1 \\
\textit{$\Delta$ Performance Gap} & & \cellcolor{gray!20}{\textit{+16.7}}  & \cellcolor{gray!20}{\textit{+18.1}}  & \cellcolor{gray!20}{\textit{+14.2}}  & \cellcolor{gray!20}{\textit{+10.7}}  & \cellcolor{gray!20}{\textit{+12.9}}  & \cellcolor{gray!20}{\textit{+16.5}}  & \cellcolor{gray!20}{\textit{+8.1}} & \cellcolor{gray!20}{\textit{+12.3}}  & \cellcolor{gray!20}{\textit{+13.7}}  \\
\midrule

Vanilla OPD~\citep{thinking2025opd} & \multirow{4}{*}{Standard} & 65.2 & 40.9 & 75.6 & \cellcolor{myblue!60}{16.7} & 15.8 & 64.3 & 55.4 & \cellcolor{myblue!60}{17.6} & 43.9 \\
OPCD~\citep{ye2026policy} &  & 66.1 & 41.6 & 75.3 & \cellcolor{myblue!60}{19.5} & 17.0 & 65.2 & 54.7 & 15.8 & 44.3 \\
ExOPD~\citep{yang2026learning} &  & \underline{68.3} & \cellcolor{myblue!60}{44.7} & 76.7 & \cellcolor{myblue!60}{18.5} & \cellcolor{myblue!60}{19.9} & \underline{68.0} & \cellcolor{myblue!60}{\underline{57.2}} & \cellcolor{myblue!60}{\underline{22.6}} & \cellcolor{myblue!60}{\underline{47.0}} \\
Uni-OPD~\citep{hou2026uni} &  & 66.5 & 42.3 & \underline{77.5} & \cellcolor{myblue!60}{\underline{20.0}} & \cellcolor{myblue!60}{\underline{22.3}} & 67.2 & \cellcolor{myblue!60}{56.1} & \cellcolor{myblue!60}{20.8} & \cellcolor{myblue!60}{46.6} \\

\midrule
SDFT~\citep{shenfeld2026self} & \multirow{3}{*}{Self} & 63.4 & 38.7 & 73.8 & \cellcolor{myblue!60}{15.0} & 15.4 & 62.6 & 53.2 & 12.1 & 41.8 \\
OPSD~\citep{zhao2026self} &  & 64.6 & 39.7 & 73.8 & \cellcolor{myblue!60}{15.2} & 14.7 & 63.1 & 54.2 & 14.5 & 42.5 \\
SDPO~\citep{hubotter2026reinforcement} &  & 65.4 & 38.8 & 74.0 & \cellcolor{myblue!60}{16.4} & 15.1 & 62.9 & 52.3 & \cellcolor{myblue!60}{17.7} & 42.8 \\

\midrule
EOPD~\citep{jin2026entropy} & \multirow{2}{*}{Adaptive} & 67.5 & \cellcolor{myblue!60}{\underline{45.7}} & 75.9 & \cellcolor{myblue!60}{17.6} & \cellcolor{myblue!60}{19.3} & 67.1 & \cellcolor{myblue!60}{56.8} & \cellcolor{myblue!60}{19.0} & \cellcolor{myblue!60}{46.1} \\
TIP~\citep{xu2026tip} &  & 67.0 & 43.1 & 74.3 & \cellcolor{myblue!60}{17.2} & \cellcolor{myblue!60}{18.7} & 65.5 & 54.5 & \cellcolor{myblue!60}{17.9} & 44.8 \\

\midrule 
\textbf{DOPD (Ours)} & Dual & \cellcolor{myblue!60}{\textbf{71.3}} & \cellcolor{myblue!60}{\textbf{49.8}} & \cellcolor{myblue!60}{\textbf{81.5}} & \cellcolor{mygreen!60}{\textbf{23.3}} & \cellcolor{mygreen!60}{\textbf{26.9}} & \cellcolor{myblue!60}{\textbf{71.0}} & \cellcolor{mygreen!60}{\textbf{60.2}} & \cellcolor{mygreen!60}{\textbf{27.1}} & \cellcolor{myblue!60}{\textbf{51.4}} \\

\bottomrule
\end{tabular}}
\label{tab:llm_main}
\end{table*}

\begin{table*}[t]
\centering
\caption{Performance comparison of our proposed DOPD with counterparts on general, visual reasoning, and visual understanding tasks. $^*$The codes of VA-OPD are not officially released, so we use the results of our reproduced version. }
\setlength{\tabcolsep}{0.9mm}
\resizebox{1\textwidth}{!}{
\begin{tabular}{l|cccccccc|c}
\toprule   
\multirow{2}{*}{\textbf{Method}} & \multicolumn{2}{c}{\textbf{General}} & \multicolumn{3}{c}{\textbf{Visual Reasoning}} & \multicolumn{3}{c|}{\textbf{Visual Understanding}} & \multirow{2}{*}{\textbf{Average}} \\ 
\cmidrule(lr){2-3} \cmidrule(lr){4-6} \cmidrule(lr){7-9}     
& \textbf{RealWorldQA} & \textbf{MMStar} &  \textbf{MathVision} & \textbf{DynaMath} & \textbf{LogicVista} & \textbf{MMMU} & \textbf{MMMU-Pro} & \textbf{VSI-Bench} &  \\  \midrule
Teacher Policy & 71.3 & 70.7 & 53.8 & 67.6 & 55.0 & 69.6 & 55.8 & 59.7 & 62.9 \\
Student Policy & 63.6 & 58.4 & 32.0 & 53.8 & 35.5 & 53.2 & 36.4 & 53.6 & 48.3 \\
\textit{$\Delta$ Performance Gap} & \cellcolor{gray!20}{\textit{+7.7}} & \cellcolor{gray!20}{\textit{+12.3}}  & \cellcolor{gray!20}{\textit{+21.8}}  & \cellcolor{gray!20}{\textit{+13.8}}  & \cellcolor{gray!20}{\textit{+19.5}}  & \cellcolor{gray!20}{\textit{+16.4}}  & \cellcolor{gray!20}{\textit{+19.4}}  & \cellcolor{gray!20}{\textit{+6.1}} & \cellcolor{gray!20}{\textit{+14.6}}  \\\midrule

Vanilla OPD~\citep{thinking2025opd} & 64.7 & 61.8 & 37.1 & 56.2 & 40.2 & 58.0 & \cellcolor{myblue!60}{46.7} & 54.1 & 52.4 \\
Uni-OPD~\citep{hou2026uni} & 65.0 & \cellcolor{myblue!60}{65.3} & \cellcolor{myblue!60}{\underline{43.0}} & \underline{58.2} & 42.5 & 59.1 & \cellcolor{myblue!60}{47.0} & 53.7 & 54.2 \\
Vision-OPD~\citep{yuan2026vision} & 66.2 & \cellcolor{myblue!60}{\underline{66.4}} & 38.0 & 57.6 & 43.1 & \cellcolor{myblue!60}{64.9} & \cellcolor{myblue!60}{52.3} & 56.1 & \cellcolor{myblue!60}{55.6} \\
VA-OPD$^*$~\citep{liu2026visual} & \underline{67.0} & \cellcolor{myblue!60}{66.2} & 38.7 & 57.7 & \underline{43.1} & \cellcolor{myblue!60}{\underline{66.1}} & \cellcolor{myblue!60}{\textbf{54.2}} & \cellcolor{myblue!60}{\underline{57.5}} & \cellcolor{myblue!60}{\underline{56.3}} \\

\midrule 
\textbf{DOPD (Ours)} & \textbf{67.4} & \cellcolor{myblue!60}{\textbf{67.2}} & \cellcolor{myblue!60}{\textbf{45.6}} & \textbf{60.5} & \cellcolor{myblue!60}{\textbf{47.7}} & \cellcolor{myblue!60}{\textbf{67.0}} & \cellcolor{myblue!60}{\underline{53.9}} & \cellcolor{myblue!60}{\textbf{57.8}} & \cellcolor{myblue!60}{\textbf{58.4}} \\

\bottomrule
\end{tabular}}
\label{tab:vlm_main}
\end{table*}

\subsubsection{Baselines} We compare our DOPD with other nine LLM-based counterparts, including three main paradigms of OPD as we discussed in Section~\ref{sec:ralated_work}: (a) standard distillation: Vanilla OPD~\citep{thinking2025opd}, OPCD~\citep{ye2026policy}, ExOPD~\citep{yang2026learning}, and Uni-OPD~\citep{hou2026uni}; (b) self distillation: SDFT~\citep{shenfeld2026self}, OPSD~\citep{zhao2026self}, and SDPO~\citep{hubotter2026reinforcement}; and (c) adaptive distillation: EOPD~\citep{jin2026entropy} and TIP~\citep{xu2026tip}.
For VLM-based methods, we benchmark DOPD with other four methods: Vanilla OPD~\citep{thinking2025opd}, Uni-OPD~\citep{hou2026uni}, Vision-OPD~\citep{yuan2026vision}, and VA-OPD~\citep{liu2026visual}. For fair comparison, we rerun all the baselines on Qwen3/Qwen3-VL models.

\subsubsection{Implementations} All experiments are conducted on 8 NVIDIA H200 141GB GPUs. During distillation, the teacher policy is frozen for stability, while the student policy is optimized by AdamW optimizer and cosine scheduler with a learning rate of $5\times10^{-6}$. The batch sizes are set to 128 and 64 for LLM and VLM with 4 rollout samples, optimizing for a maximum of 200 and 300 steps, respectively.
The K is set to 128 for Top-K distillation, and $\beta_{w}$ and $\beta_{l}$ are 0.3 and 0.6 to regulate the strength of distillation.

\subsection{Main Results}

\subsubsection{Distillation Performance}
As the main LLM-based OPD results reported in Table~\ref{tab:llm_main}, DOPD substantially narrows the performance gap between the student and teacher policies, with a gain of 12.3 points and an 89.8\% recovery of the original teacher-student gap. Notably, due to the introduction of privileged information that increases the upper limit of distillation, DOPD not only approaches the teacher policy on average, but also surpasses the teacher on four challenging benchmarks, especially on reasoning and coding tasks. 
Compared with standard (\textit{i.e.}, strong-to-weak) and adaptive distillation counterparts, DOPD consistently achieves the best performance across all eight benchmarks and improves over the three strongest baselines, ExOPD~\citep{yang2026learning}/Uni-OPD~\citep{hou2026uni}/EOPD~\citep{jin2026entropy} by 4.4/4.8/5.3 points on average, respectively. 
Self-distillation baselines provide relatively modest improvements,  suggesting that existing methods relying solely on the self-distillation of the student is possibly insufficient for closing the teacher-student gap.

We further validate the effectiveness on VLM-based OPD, as listed in Table~\ref{tab:vlm_main}. Specifically, our proposed DOPD again brings a substantial improvement over the student policy by  a 10.1-point absolute gain and a 69.2\% recovery of the teacher-student gap.  Compared with existing VLM-oriented OPD baselines, DOPD achieves the best average performance, outperforming Vanilla OPD~\citep{thinking2025opd}, and other three baselines, Uni-OPD~\citep{hou2026uni}, Vision-OPD~\citep{yuan2026vision}, and VA-OPD~\citep{liu2026visual}, by 6.0 and 4.2/2.8/2.1 points, respectively. 
It is worth mentioning that all methods, including ours, have shown more significant improvements in visual understanding than reasoning and other visual tasks, which may be related to the distillation paradigm of the visual center, mainly distilling accurate and grounded focus from teacher on the visual evidence.

These results demonstrate that advantage-aware dual distillation is more effective than either static teacher imitation, self-refinement, or single-sided adaptive weighting, indicating that DOPD transfers not only surface-level output preferences but also more essential ability from teacher policy. 
In addition, beyond text-only distillation, the proposed paradigm also provides robust and consistent gains for vision ability distillation.

\begin{table*}[t]
\centering
\caption{Generalization comparison of our proposed DOPD and Vanilla OPD based on five pairs of teacher-student models, including Qwen3-8B/4B/1.7B $\rightarrow$ Qwen3-0.6B, and Qwen3-8B/4B $\rightarrow$ Qwen3-1.7B.}
\setlength{\tabcolsep}{0.9mm}
\resizebox{1\textwidth}{!}{
\begin{tabular}{l|l|cccccccc|c}
\toprule   
\multirow{2}{*}{\textbf{Model Pair}} & \multirow{2}{*}{\textbf{Method}} & \multicolumn{2}{c}{\textbf{General}} & \multicolumn{4}{c}{\textbf{Reasoning}} & \multicolumn{2}{c|}{\textbf{Coding}} & \multirow{2}{*}{\textbf{Average}} \\ 
\cmidrule(lr){3-4} \cmidrule(lr){5-8} \cmidrule(lr){9-10}     
& & \textbf{C-Eval} & \textbf{LiveBench} &  \textbf{MATH500} & \textbf{AIME25} & \textbf{ZebraLogic} & \textbf{AutoLogi} & \textbf{BFCLv3}  & \textbf{LCBv5} &  \\  \midrule

 & \multicolumn{1}{c|}{Qwen3-8B} & 77.1 & 53.5 & 86.9 & 20.2 & 25.0 & 76.3 & 60.0 & 23.6 & 52.8 \\
Base & \multicolumn{1}{c|}{Qwen3-4B} & 72.2 & 48.3 & 84.6 & 18.9 & 35.0 & 75.8 & 57.4 & 21.3 & 51.6 \\
Model & \multicolumn{1}{c|}{Qwen3-1.7B} & 60.4 & 35.4 & 72.7 & 9.5 & 12.1 & 59.8 & 51.9 & 11.3 & 39.1 \\
& \multicolumn{1}{c|}{Qwen3-0.6B} & 42.0 & 21.8 & 55.2 & 2.0 & 4.1 & 37.2 & 44.0 & 3.5 & 26.2 \\
\midrule

Qwen3-8B & \textit{$\Delta$ Performance Gap} & \cellcolor{gray!20}{\textit{+35.1}} & \cellcolor{gray!20}{\textit{+31.7}} & \cellcolor{gray!20}{\textit{+31.7}} & \cellcolor{gray!20}{\textit{+18.2}} & \cellcolor{gray!20}{\textit{+20.9}} & \cellcolor{gray!20}{\textit{+39.1}} & \cellcolor{gray!20}{\textit{+16.0}} & \cellcolor{gray!20}{\textit{+20.1}} & \cellcolor{gray!20}{\textit{+26.6}}  \\
\multicolumn{1}{c|}{$\downarrow$} & Vanilla OPD & 44.8 & 24.5 & 59.3 & 4.6 & 5.9 & 46.5 & 46.1 & 6.1 & 29.7 \\
Qwen3-0.6B & \textbf{DOPD (Ours)} & \textbf{56.7} & \textbf{35.5} & \cellcolor{myblue!60}{\textbf{70.9}} & \cellcolor{myblue!60}{\textbf{17.0}} & \cellcolor{myblue!60}{\textbf{19.6}} & \textbf{54.3} & \cellcolor{myblue!60}{\textbf{55.5}} & \textbf{12.7} & \cellcolor{myblue!60}{\textbf{40.3}} \\
\midrule

Qwen3-8B & \textit{$\Delta$ Performance Gap} &   \cellcolor{gray!20}{\textit{+16.7}}  & \cellcolor{gray!20}{\textit{+18.1}}  & \cellcolor{gray!20}{\textit{+14.2}}  & \cellcolor{gray!20}{\textit{+10.7}}  & \cellcolor{gray!20}{\textit{+12.9}}  & \cellcolor{gray!20}{\textit{+16.5}}  & \cellcolor{gray!20}{\textit{+8.1}} & \cellcolor{gray!20}{\textit{+12.3}}  & \cellcolor{gray!20}{\textit{+13.7}} \\
\multicolumn{1}{c|}{$\downarrow$} & Vanilla OPD & 65.2 & 40.9 & 75.6 & \cellcolor{myblue!60}{16.7} & 15.8 & 64.3 & 55.4 & \cellcolor{myblue!60}{17.6} & 43.9 \\
Qwen3-1.7B & \textbf{DOPD (Ours)} & \cellcolor{myblue!60}{\textbf{71.3}} & \cellcolor{myblue!60}{\textbf{49.8}} & \cellcolor{myblue!60}{\textbf{81.5}} & \cellcolor{mygreen!60}{\textbf{23.3}} & \cellcolor{mygreen!60}{\textbf{26.9}} & \cellcolor{myblue!60}{\textbf{71.0}} & \cellcolor{mygreen!60}{\textbf{60.2}} & \cellcolor{mygreen!60}{\textbf{27.1}} & \cellcolor{myblue!60}{\textbf{51.4}}  \\
\midrule

Qwen3-4B & \textit{$\Delta$ Performance Gap} & \cellcolor{gray!20}{\textit{+30.2}} & \cellcolor{gray!20}{\textit{+26.5}} & \cellcolor{gray!20}{\textit{+29.4}} & \cellcolor{gray!20}{\textit{+16.9}} & \cellcolor{gray!20}{\textit{+30.9}} & \cellcolor{gray!20}{\textit{+38.6}} & \cellcolor{gray!20}{\textit{+13.4}} & \cellcolor{gray!20}{\textit{+17.8}} & \cellcolor{gray!20}{\textit{+25.4}} \\
\multicolumn{1}{c|}{$\downarrow$} & Vanilla OPD & 44.8 & 24.0 & 60.9 & 7.4 & 7.9 & 47.4 & 46.5 & 8.6 & 30.9 \\
Qwen3-0.6B & \textbf{DOPD (Ours)} & \textbf{54.0} & \textbf{33.8} & \cellcolor{myblue!60}{\textbf{72.7}} & \cellcolor{myblue!60}{\textbf{16.3}} & \cellcolor{myblue!60}{\textbf{25.1}} & \textbf{51.6} & \cellcolor{myblue!60}{\textbf{52.2}} & \cellcolor{myblue!60}{\textbf{13.7}} & \cellcolor{myblue!60}{\textbf{39.9}} \\
\midrule

Qwen3-4B & \textit{$\Delta$ Performance Gap} & \cellcolor{gray!20}{\textit{+11.8}} & \cellcolor{gray!20}{\textit{+12.9}} & \cellcolor{gray!20}{\textit{+11.9}} & \cellcolor{gray!20}{\textit{+9.4}} & \cellcolor{gray!20}{\textit{+22.9}} & \cellcolor{gray!20}{\textit{+16.0}} & \cellcolor{gray!20}{\textit{+5.5}} & \cellcolor{gray!20}{\textit{+10.0}} & \cellcolor{gray!20}{\textit{+12.5}} \\
\multicolumn{1}{c|}{$\downarrow$} & Vanilla OPD & 65.3 & 41.0 & 76.3 & \cellcolor{myblue!60}{17.1} & 16.8 & 65.6 & 53.5 & \cellcolor{myblue!60}{16.6} & 44.0 \\
Qwen3-1.7B & \textbf{DOPD (Ours)} & \cellcolor{myblue!60}{\textbf{69.9}} & \cellcolor{myblue!60}{\textbf{48.1}} & \cellcolor{myblue!60}{\textbf{80.2}} & \cellcolor{mygreen!60}{\textbf{22.0}} & \cellcolor{myblue!60}{\textbf{29.1}} & \cellcolor{myblue!60}{\textbf{70.8}} & \cellcolor{myblue!60}{\textbf{57.2}} & \cellcolor{mygreen!60}{\textbf{24.4}} & \cellcolor{myblue!60}{\textbf{50.2}} \\
\midrule

Qwen3-1.7B & \textit{$\Delta$ Performance Gap} & \cellcolor{gray!20}{\textit{+18.4}} & \cellcolor{gray!20}{\textit{+13.6}} & \cellcolor{gray!20}{\textit{+17.5}} & \cellcolor{gray!20}{\textit{+7.5}} & \cellcolor{gray!20}{\textit{+8.0}} & \cellcolor{gray!20}{\textit{+22.6}} & \cellcolor{gray!20}{\textit{+7.9}} & \cellcolor{gray!20}{\textit{+7.8}} & \cellcolor{gray!20}{\textit{+12.9}}
\\
\multicolumn{1}{c|}{$\downarrow$} & Vanilla OPD & 47.3 & \cellcolor{myblue!60}{29.5} & 60.1 & \cellcolor{myblue!60}{7.8} & \cellcolor{myblue!60}{9.9} & 43.3 & 47.1 & \cellcolor{myblue!60}{8.2} & 31.7 \\
Qwen3-0.6B& \textbf{DOPD (Ours)} & \cellcolor{myblue!60}{\textbf{55.6}} & \cellcolor{mygreen!60}{\textbf{35.6}} & \cellcolor{myblue!60}{\textbf{66.7}} & \cellcolor{mygreen!60}{\textbf{14.6}} & \cellcolor{mygreen!60}{\textbf{16.4}} & \cellcolor{myblue!60}{\textbf{49.4}} & \cellcolor{mygreen!60}{\textbf{53.0}} & \cellcolor{mygreen!60}{\textbf{13.7}} & \cellcolor{myblue!60}{\textbf{38.1}} \\

\bottomrule
\end{tabular}}
\label{tab:llm_more}

\end{table*}

\begin{figure*}[t]
    \begin{subfigure}{0.48\linewidth}
        \centering
        \includegraphics[width=\linewidth]{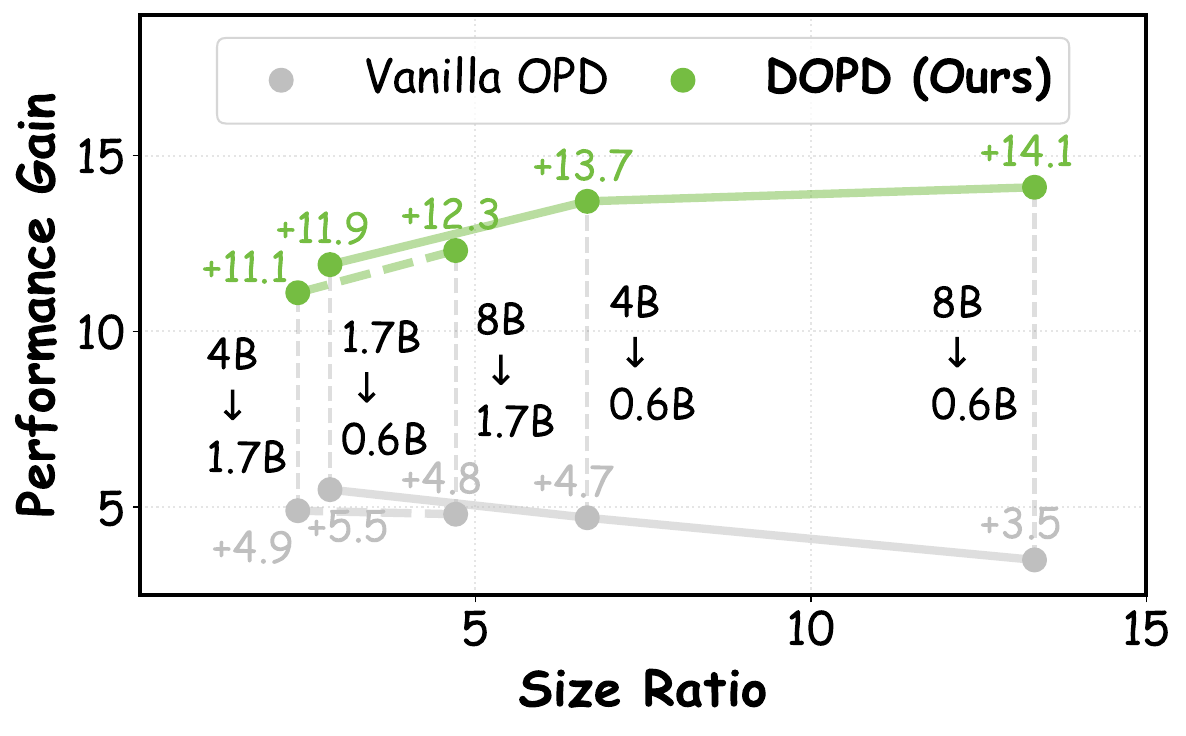}
        \caption{Performance Gain \textit{vs.} Teacher-student Size Ratio}
        \label{fig:more-a}
    \end{subfigure}
    \hfill
    \begin{subfigure}{0.48\linewidth}
        \centering
        \includegraphics[width=\linewidth]{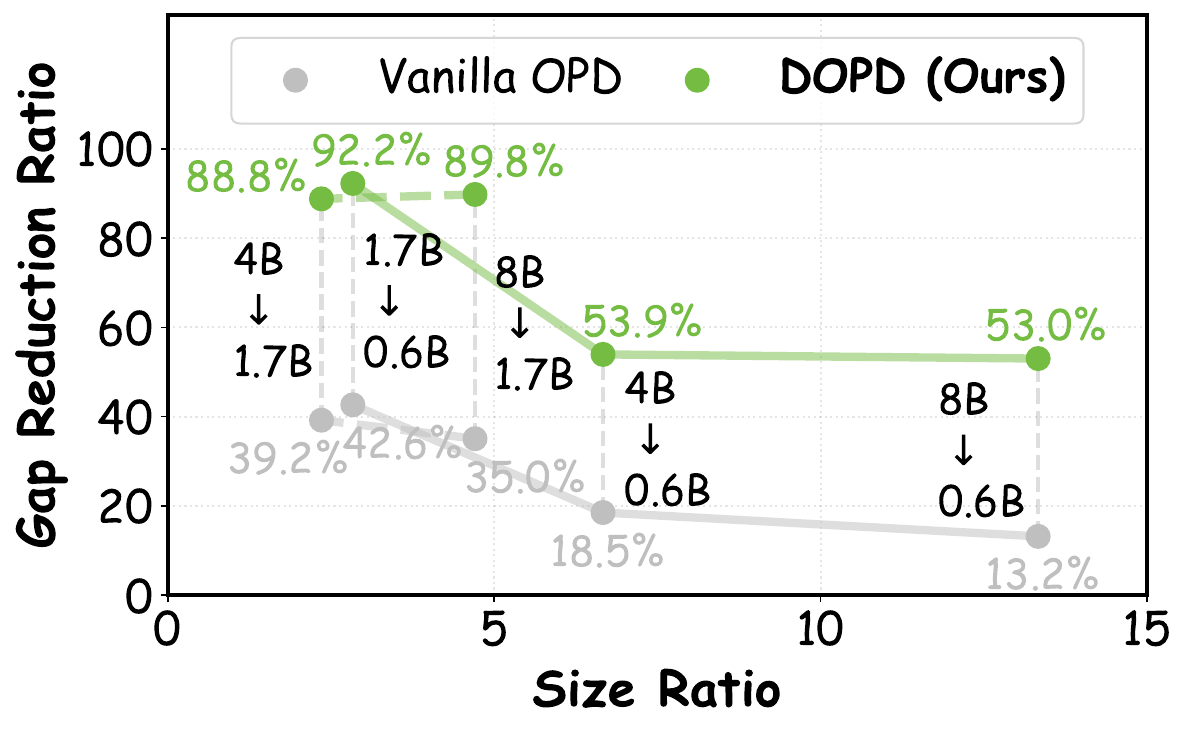} 
        \caption{Gap Reduction \textit{vs.} Teacher-student Size Ratio}
        \label{fig:more-b}
    \end{subfigure}
    
    \caption{Scalability comparison of proposed DOPD and Vanilla OPD on \textbf{(a)} performance gain and \textbf{(b)} teacher-student gap reduction ratio. Here, the solid and dashed lines represent the 0.6B and 1.7B student policy, respectively. }
    \label{fig:llm_more}
\end{figure*}

\subsubsection{Robustness \& Scalability}

To further examine whether DOPD generalizes across different teacher-student scales, we conduct experiments with five teacher-student pairs . 
Table~\ref{tab:llm_more} shows that  DOPD consistently outperforms Vanilla OPD~\citep{thinking2025opd} on every model pair, demonstrating that its effectiveness is not tied to a specific teacher or student size. 
Our proposed method achieves consistent and significant performance improvements, averaging 11.1–14.1 points across all pairs, a two- to over three-fold improvement relative to Vanilla OPD.

More importantly, our method remains robust as the teacher-student size ratio increases. 
As mentioned in previous studies~\citep{thinking2025opd,li2026rethinking,qin2026near}, a larger size ratio implies greater initial distribution inconsistency between teachers and students, which may lead to suboptimal distillation effects.
For instance, in the largest scale-mismatch setting, \textit{i.e.}, Qwen3-8B $\rightarrow$ Qwen3-0.6B, Vanilla OPD only reaches a 3.5-point gain;  In contrast,  DOPD achieves a 14.1-point gain and recovers 53.0\% of the teacher-student gap.
Similar trends can be observed for Qwen3-4B $\rightarrow$ Qwen3-0.6B.
As illustrated in Figure~\ref{fig:more-a}, when the teacher model has larger parameters, and stronger capabilities, the performance improvement of Vanilla OPD actually decreases, suggesting that naive imitation becomes less effective when the capacity mismatch is large. 
By contrast, DOPD maintains gradually increasing gains across these settings.
Furthermore, as reported in Figure~\ref{fig:more-b}, although the gap reduction inevitably decreases as the size ratio increases, due to the larger initial teacher-student gap and the limitations of the ability limit  of student model, our model still effectively alleviates this trend. 
These results indicate that DOPD provides a more scalable and reliable distillation mechanism, especially when transferring policies from substantially larger teachers to compact students.

\begin{figure}[t]
    \begin{subfigure}{0.60\linewidth}
        \centering
        \includegraphics[width=\linewidth]{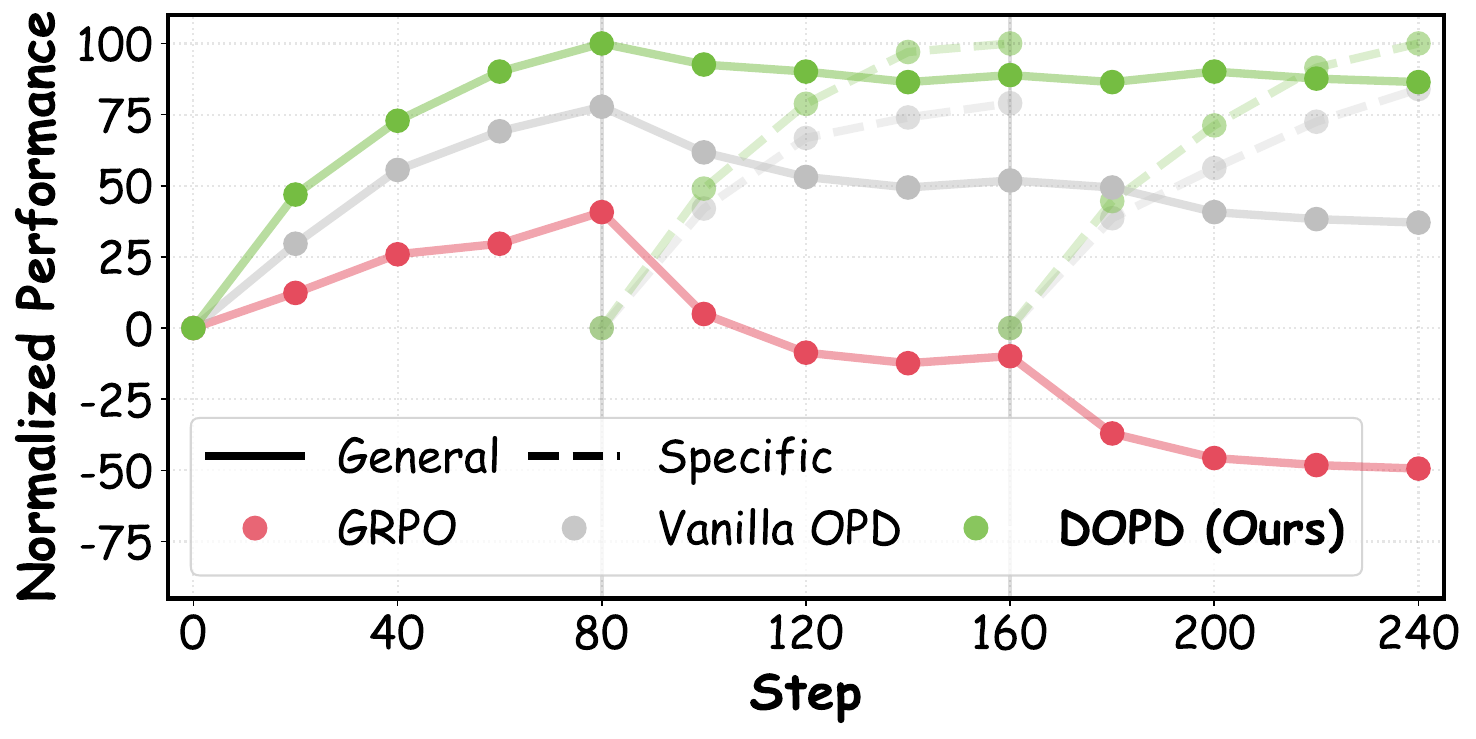}
        \caption{Normalized Performance on Three-stage Continual Learning}
        \label{fig:continual}
    \end{subfigure}
    \hfill
    \begin{subfigure}{0.3\linewidth}
        \centering
        \includegraphics[width=\linewidth]{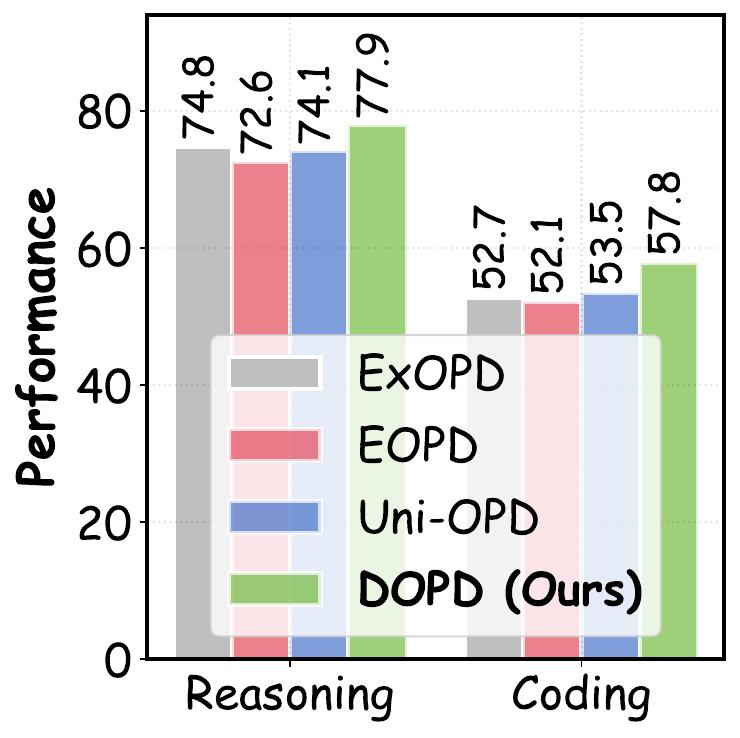} 
        \caption{Out-of-distribution Evaluation}
        \label{fig:ood}
    \end{subfigure}
    
    \caption{Comparison of proposed DOPD and Vanilla OPD on \textbf{(a)} continual learning, where we conduct a three-stage continual learning with general, reasoning, and coding training sub-datasets sequentially. The solid and dashed lines denote the results on general benchmark (LiveBench) and corresponding specific benchmarks (MATH500 and BFCLv3); and \textbf{(b)} out-of-distribution tasks, where we optimize the student policy on coding or reasoning dataset, but evaluated on another out-of-domain benchmarks (MATH500 and BFCLv3).}
    \label{fig:advanced}
\end{figure}

\subsubsection{Continual Learning Evaluation}
OPD has been demonstrated to yield superior performance in continual learning, mitigating the catastrophic forgetting~\citep{shenfeld2026self,hubotter2026reinforcement} inherent to several prevalent post-training paradigms, \textit{e.g.}, SFT and GRPO~\citep{shao2024deepseekmath}. Thus, we perform a three-stage experiment to evaluate the continual learning performance, where in the first stage only add general training data, while use reasoning and coding data in the next two stages.

Figure~\ref{fig:continual} indicates OPD-based paradigms have significantly better sustained learning performance and less forgetting, and our DOPD further optimizes this advantage. Specifically, it supports steady and effective capability accumulation: performance improves consistently on each newly introduced data domain, with only tiny performance degradation on previously acquired domains. This finding validates that DOPD enables authentic continual learning, where a single model can incrementally gain multiple capabilities instead of relying on simple capability concatenation or overwriting.

\begin{figure}[t]
    \begin{subfigure}{0.48\linewidth}
        \centering
        \includegraphics[width=\linewidth]{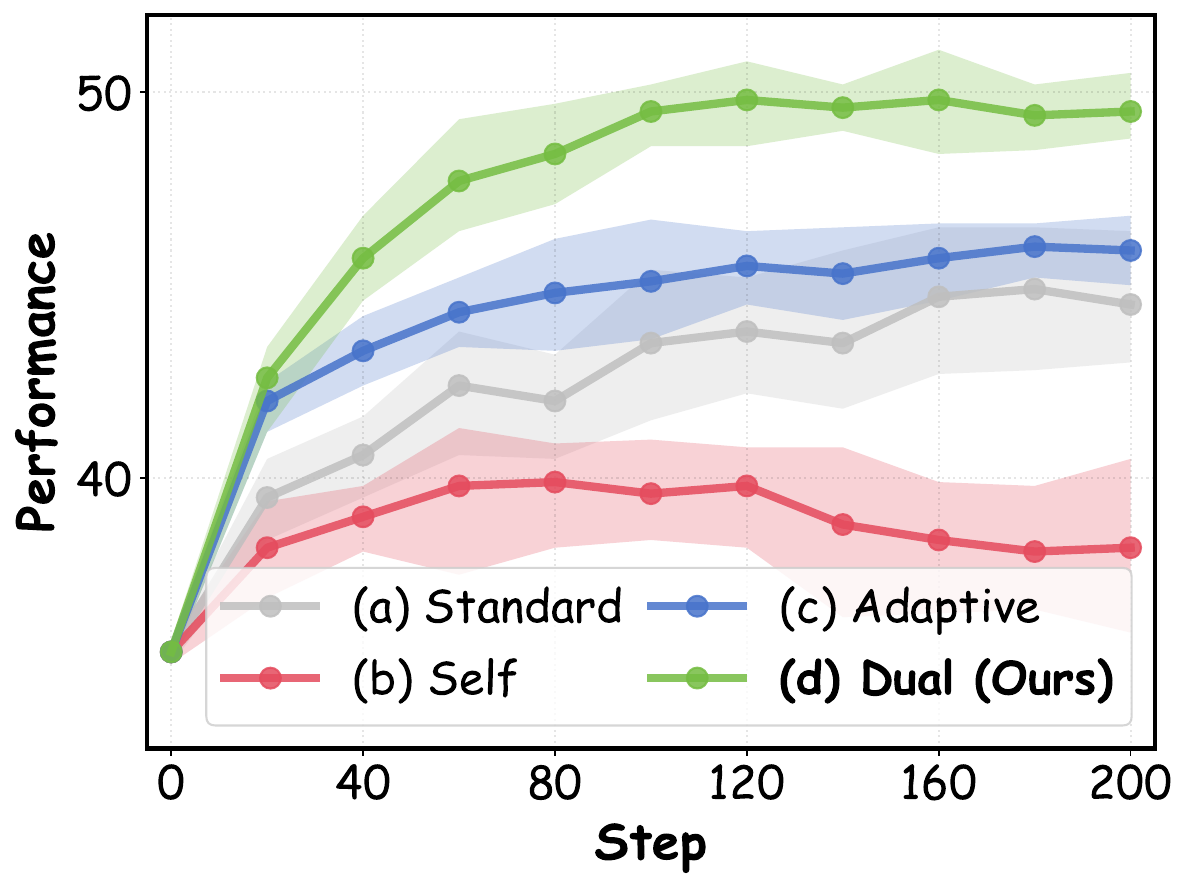}
        \caption{Performance \textit{vs.} Training Step}
        \label{fig:com-performance}
    \end{subfigure}
    \hfill
    \begin{subfigure}{0.48\linewidth}
        \centering
        \includegraphics[width=\linewidth]{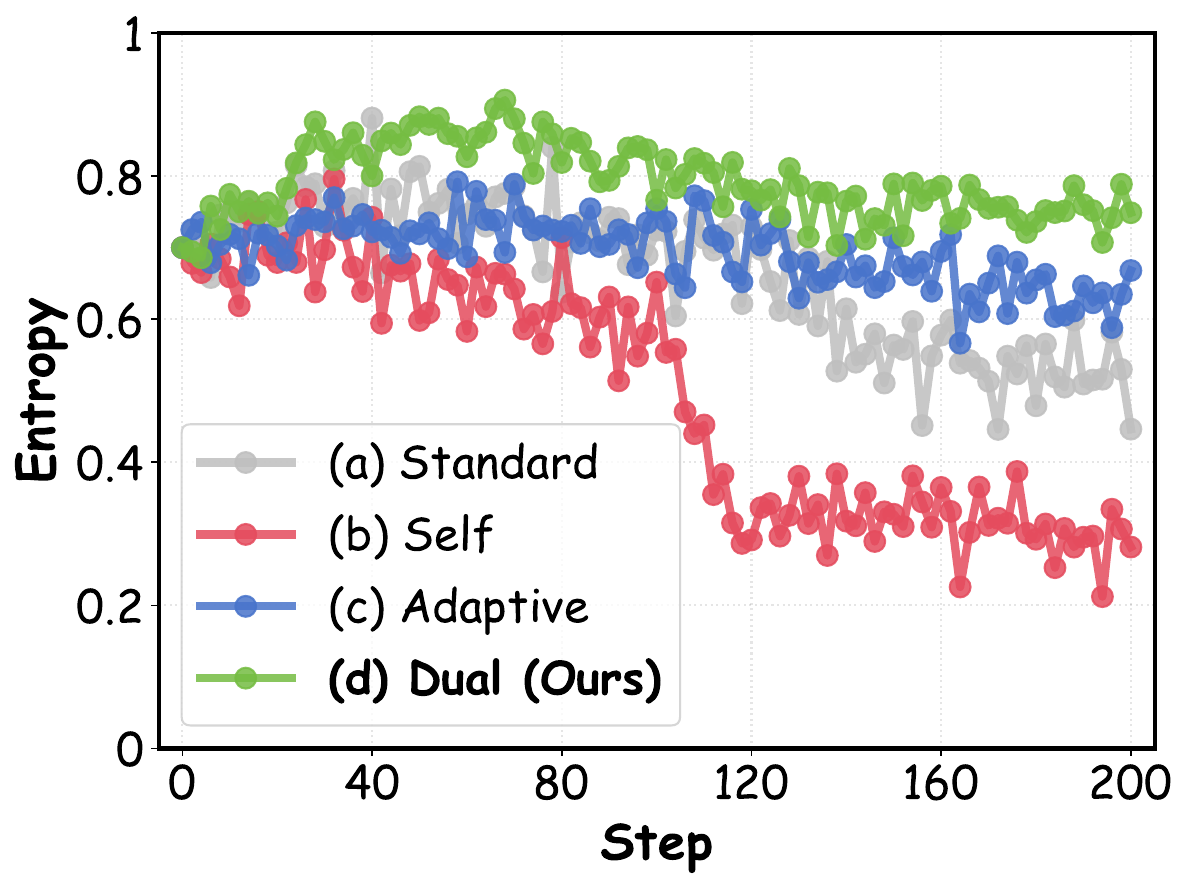} 
        \caption{Entropy \textit{vs.} Training Step}
        \label{fig:com-entropy}
    \end{subfigure}
    \caption{Training stability comparison of proposed DOPD and representative baselines, reporting the \textbf{(a)} performance and \textbf{(b)} entropy trends over training steps on LiveBench.}
    \label{fig:training}
\end{figure}

\subsubsection{Out-of-distribution Evaluation}
We further evaluate the out-of-distribution generalization. Specifically, we optimize models on either the coding or reasoning training set separately, and assess their performance on the other unseen out-of-domain tasks. For comparative analysis, we select three best-performing baselines: ExOPD~\citep{yang2026learning}, Uni-OPD~\citep{hou2026uni}, and EOPD~\citep{jin2026entropy}. As demonstrated in Figure~\ref{fig:ood}, our proposed DOPD outperforms the second-best counterparts by 3.1 and 4.3 points respectively, showcasing superior cross-domain generalization.

\subsection{Additional Analyses}
\subsubsection{Training Stability}
To further assess training stability, we benchmark our method against the best-performing baselines from three distinct distillation paradigms: ExOPD~\citep{yang2026learning} for standard distillation, SDPO~\citep{hubotter2026reinforcement} for self-distillation, and EOPD~\citep{jin2026entropy} for adaptive distillation.
As depicted in Figure~\ref{fig:com-performance}, our method consistently delivers stable and superior performance throughout the entire training process, coupled with higher distillation efficiency. Compared with the three competing paradigms, our method surpasses their step-200 performance as early as step-80.
As shown in Figure~\ref{fig:com-entropy}, our method maintains a healthy entropy trajectory: it rises modestly in the early training stage, followed by a gradual decline, and converges to a steady state after step-110. This pattern reflects that the model undergoes stable learning with well-calibrated exploration.
Notably, we observe that the self-distillation paradigm encounters entropy collapse around step-95, alongside a subsequent drop in performance. This degradation is likely attributable to the insufficient and overly homogeneous supervision signals inherent to this paradigm, which render the learned distribution deficient in necessary exploration.
Collectively, these results corroborate that our proposed method achieves superior performance gains in a stable and efficient manner throughout the distillation process.

\begin{table}[t]

\begin{minipage}[b]{0.47\textwidth}
\centering
\caption{Comparison of various LLM-based privileged information incorporation.}
\setlength{\tabcolsep}{0.9mm}
\resizebox{1\textwidth}{!}{
\begin{tabular}{l|cc}
\toprule
\multirow{1}{*}{\textbf{Privileged Input}} & \textbf{C-Eval} & \textbf{LiveBench} \\  \midrule
Final Answer & 59.5 & 36.7 \\
Step-wise Hints with Execution & 63.1 & 38.9 \\
\textbf{Step-wise Hints without Execution} & \textbf{71.3} & \textbf{49.8} \\
Summarized Hints & 65.8 & 43.6 \\
No Privileged Input & 63.0 & 39.4 \\
\bottomrule
\end{tabular}}
\label{tab:pi_com_llm}
\end{minipage}
\hfill
\begin{minipage}[b]{0.49\textwidth}
\centering
\caption{Comparison of various VLM-based privileged information incorporation.}
\setlength{\tabcolsep}{0.9mm}
\resizebox{1\textwidth}{!}{
\begin{tabular}{l|cc}
\toprule
\multirow{1}{*}{\textbf{Privileged Input}} & \textbf{RealWorldQA} & \textbf{MMStar} \\  \midrule
Final Answer & 64.6 & 63.2 \\
Bounding Box with Caption & 65.3 & 66.9 \\
\textbf{Bounding Box with Object Label} & \textbf{67.4} & \textbf{67.2} \\ 
Caption & 64.8 & 65.6 \\
No Privileged Input & 63.2 & 60.0 \\
\bottomrule
\end{tabular}}
\label{tab:pi_com_vlm}
\end{minipage}

\end{table}

\begin{figure}[t]
    \vspace{10pt}
    \centering
    \includegraphics[width=0.9\linewidth]{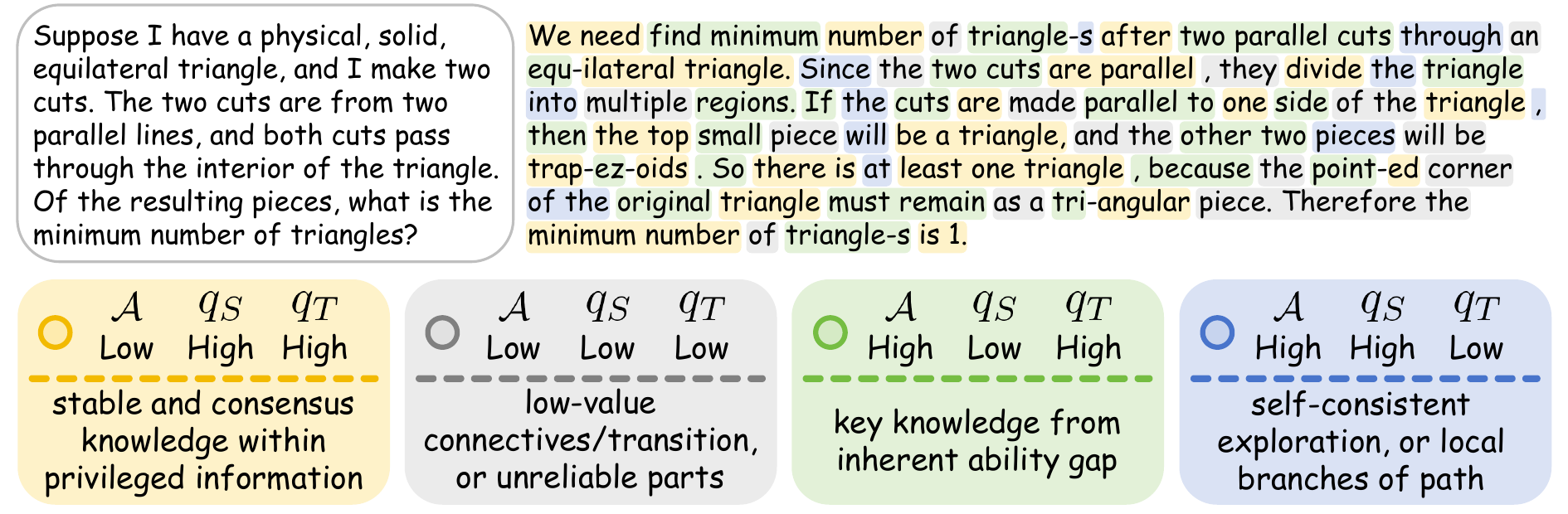}
    \caption{Token-level visualization of the four token types, where each token is colored based on their privilege advantage gap $\mathcal{A}$ and predicted probabilities of teacher $q_{T}$ and student $q_{S}$ policies.}
    \label{fig:token_demo}
\end{figure}

\subsubsection{Privileged Information Analysis}
\label{sec:pi_analysis}

To evaluate the impacts of distinct privileged information injection strategies, we conduct comparative experiments to benchmark the performance of five different privileged information formulations: final answer, step-wise hints with detailed execution process, step-wise hints without execution,  simplest summarized hints, and no privileged input for LLM-based distillation, and final answer, bounding box with descriptive caption, bounding box with object label, caption, and no privileged information for VLM-based task.
As summarized in Table~\ref{tab:pi_com_llm} and \ref{tab:pi_com_vlm}, directly providing ground-truth answers incurs the most severe information gap. The student model can only rigidly overfit to the given answers, which induces potential shortcut learning and performance degradation, even underperforming the baseline without any privileged information.
In contrast, providing only step-level high-level hints without detailed execution steps yields the largest LLM distillation gains of 8.3 and 10.4 points respectively. Meanwhile, providing bounding boxes paired with corresponding object labels proves to be the most suitable privileged information modality for VLM, bringing 4.2 and 7.2 points of improvement over the baseline.
Notably, the efficacy of privileged information does not lie in the correctness of the final answers, but rather in its ability to deliver capability-oriented guidance to the student model, consistent with our previous discussion in Section~\ref{sec:background}.

\subsubsection{Token Analysis}
As detailed and analyzed in Section~\ref{sec:background} and \ref{sec:dopd}, we first compute the privilege advantage gap $\mathcal{A}$ and the predicted probabilities $q_{T/S}$ of both the teacher and student policies for each token, based on which we categorize each token into distinct classes.
To intuitively characterize the functional roles of different token types during distillation, Figure~\ref{fig:token_demo} visualizes the distribution of token categories within a real trajectory.
Among low-gap tokens, those with both high probabilities typically correspond to stable and consensus knowledge within privileged information, whereas tokens with both low probabilities are mostly connectives, transitions or unreliable segments with little valid information.
Among high-gap tokens, tokens with high teacher probability but low student probability generally represent key knowledge arising from the inherent privilege-conditioned ability gap, while tokens with high student probability but low teacher probability likely reflect self-consistent or local branches of exploration.
This token distribution pattern aligns well with our proposed token-level differentiated distillation strategy, enabling targeted and efficient distillation for tokens with distinct functional roles.

To further quantitatively dissect the contributions of individual token types and the adaptive advantage-aware dual distillation mechanism to our proposed method, we conduct token-level ablation analysis.
Each variant performs distillation with signals from only one or combinations of token types, utilizing JS divergence on Top-K tokens. The setting without adaptive distillation corresponds to a baseline where all tokens receive identical distillation weights and strategies, with no token-wise differentiation.
As listed in Table~\ref{tab:token}, using exclusively tokens with high teacher probability and low student probability already outperforms the equal-distillation setup using all four token types (equivalent to Vanilla OPD) by 4.6 points.
However, naively adding the other three token types under an equal distillation scheme yields only marginal performance gains, and may even cause performance degradation.
In contrast, equipping the framework with the adaptive distillation mechanism allows for adjustment of token-level distillation intensity, supervision granularity, and distillation content. These designed patterns render the distillation process more efficient and stable, delivering an overall improvement of over 8 points than equal distillation, when all four token types are leveraged.

\begin{table}[t]
\begin{minipage}[t]{0.46\textwidth}
\centering
\caption{Effectiveness of individual or combinations of four tokens, and adaptive mechanism on LiveBench.}
\setlength{\tabcolsep}{0.9mm}
\resizebox{1\textwidth}{!}{
\begin{tabular}{c|ccc|c|ccc}
\toprule
\tokentypethree & \tokentypeone & \tokentypetwo & \tokentypefour & \textbf{Adaptive} & \textbf{Step-40} & \textbf{Step-80} & \textbf{Step-160} \\ \midrule
\cmark & \xmark & \xmark & \xmark & \xmark & 41.0 & 43.7 & 45.3 \\
\midrule
\cmark & \cmark & \xmark & \xmark & \xmark & 39.9 & 42.9 & 45.0 \\
\cmark & \xmark & \cmark & \xmark & \xmark & 38.6 & 41.0 & 42.7 \\
\cmark & \xmark & \xmark & \cmark & \xmark & 39.2 & 42.7 & 44.5 \\
\cmark & \cmark & \cmark & \cmark & \xmark & 38.4 & 40.9 & 41.3 \\
\midrule
\cmark & \cmark & \xmark & \xmark & \cmark & 41.0 & 45.4 & 47.9 \\
\cmark & \xmark & \cmark & \xmark & \cmark & 39.7 & 42.3 & 44.8 \\
\cmark & \xmark & \xmark & \cmark & \cmark & 40.1 & 44.9 & 47.6 \\
\midrule
\cmark & \cmark & \cmark & \cmark & \cmark & \textbf{45.7} & \textbf{48.4} & \textbf{49.8} \\
\bottomrule
\end{tabular}}
\label{tab:token}
\end{minipage}
\hfill
\begin{minipage}[t]{0.49\textwidth}
\centering
\caption{Impact of different divergence objectives and strategies on LiveBench.}
\setlength{\tabcolsep}{0.9mm}
\resizebox{1\textwidth}{!}{
\begin{tabular}{l|l|ccc}
\toprule
\multirow{1}{*}{\textbf{Objective}} & \multirow{1}{*}{\textbf{Strategy}} & \textbf{Step-40} & \textbf{Step-80} & \textbf{Step-160} \\  \midrule

\multirow{3}{*}{Forward KL}
& Sampled Token & 37.2 & 39.8 & 41.1 \\
& Top-K Tokens & 38.5 & 40.2 & 41.3 \\
& Full Vocabulary & 38.0 & 40.1 & 41.5 \\
\midrule

\multirow{3}{*}{Reverse KL}
& Sampled Token & 37.0 & 38.4 & 40.6 \\
& Top-K Tokens & 37.8 & 39.0 & 40.8 \\
& Full Vocabulary & 38.5 & 40.6 & 41.9 \\
\midrule

\multirow{3}{*}{JS Divergence}
& Sampled Token & 37.3 & 39.2 & 41.0 \\
& Top-K Tokens & 38.4 & 40.9 & 41.3 \\
& Full Vocabulary & 39.2 & 41.6 & 42.5 \\


\bottomrule
\end{tabular}}
\label{tab:kl}
\end{minipage}
\hfill
\begin{minipage}[b]{0.53\textwidth}
\centering
\caption{Ablation study on our DOPD, covering the main designs of advantage-aware dual distillation.}
\setlength{\tabcolsep}{0.9mm}
\resizebox{1\textwidth}{!}{
\begin{tabular}{l|cc}
\toprule
\multirow{1}{*}{\textbf{Variant}} & \textbf{C-Eval} & \textbf{LiveBench} \\  \midrule
\textit{w/o} Privileged Input & 63.6 & 38.3 \\
\textit{w/o} Distillation from Student Policy & 70.4 & 47.9 \\
\textit{w/o} Distillation from Teacher Policy & 65.9 & 41.2 \\
\midrule
\textit{w/o} Advantage-aware Distillation & 67.6 & 41.3 \\
\textit{w/o} Adaptive Divergence Objectives & 70.0 & 46.7 \\
\textit{w/o} Adaptive Divergence Strategies & 70.8 & 46.1 \\
\midrule
\textbf{DOPD (Ours)} & \textbf{71.3} & \textbf{49.8} \\
\bottomrule
\end{tabular}}
\label{tab:ablation}
\hfill
\end{minipage}
\begin{minipage}[b]{0.45\textwidth}
    \centering
    \includegraphics[width=\linewidth]{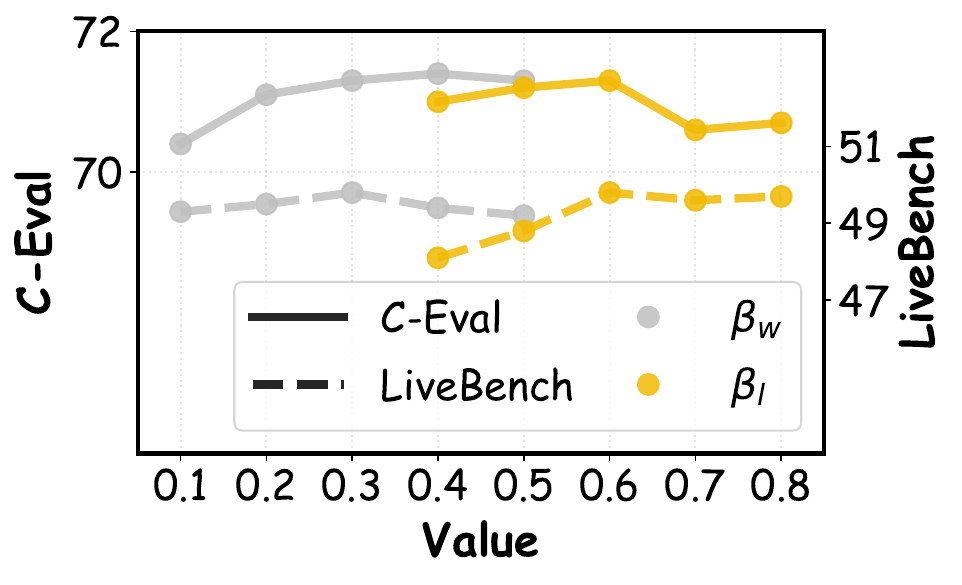}
    \vspace{-20pt}
    \captionof{figure}{Sensitivity study on the intensity coefficient of weak $\beta_{w}$ and light $\beta_{l}$ distillation.}
    \label{fig:sensitive}
\end{minipage}

\end{table}

\subsubsection{Divergence Analysis}
To further investigate the impacts of different divergence objectives (forward KL, reverse KL, and JS divergence) and strategies (sampled token, Top-K tokens, and full vocabulary), all introduced in Section~\ref{sec:dopd}, we conduct additional comparative experiments.
To isolate the effects of other factors, we apply equal distillation across all tokens. Table~\ref{tab:kl} summarizes how these design choices shape final distillation performance and efficiency.
Specifically, as the alignment scope expands from sampled to Top-K tokens and further to full vocabulary, performance improves progressively, yet inevitably incurs higher computational memory overhead.
Furthermore, in contrast to findings reported in some prior works~\citep{zhao2026self,jin2026entropy}, JS divergence delivers relatively superior performance than forward or reverse KL methods under our settings.
Collectively, these results illustrate the inherent trade-off across different divergence configurations, providing empirical justification for our differentiated distillation paradigms.

\subsubsection{Sensitivity \& Ablation Studies}
As illustrated in Figure~\ref{fig:sensitive}, we conduct an analysis focusing on the distillation intensity assigned to different token categories. We observe that setting $\beta_{w}=0.3$ and $\beta_{l}=0.6$ strikes a favorable trade-off across token-wise distillation strengths: it amplifies the contribution of critical tokens while preserving the auxiliary role of other tokens in stabilizing and providing additional optimization signals.

Furthermore, to further disentangle the contributions of individual design components in our framework, we conduct ablation studies on two core elements: the sources of distillation signals and divergence-based designs.
As presented in Table~\ref{tab:ablation}, privileged input is indispensable to our paradigm, as it directly underpins the advantage-aware calculation of our approach. Signals derived from the teacher policy serve as the primary driver of performance gains, while the student policy also fulfills an irreplaceable role throughout the distillation process. In addition, our token-wise divergence design tailored for distinct token categories is empirically validated to be effective.

\section{Conclusion}
In this work, we revisit OPD under privileged contexts and identify fundamental limitations: the apparent superiority of a privileged teacher does not always correspond to transferable capability, but may instead arise from information asymmetry, and these supervision signals are not evenly distributed across tokens.
Motivated by these observations, we propose DOPD, an advantage-aware dual on-policy distillation framework that adaptively routes token-level supervision between teacher-driven capability transfer and auxiliary self-optimization from the student. 
By leveraging the privilege advantage gap and relative token probabilities, DOPD selectively applies strong full-vocabulary teacher distillation to capability-bearing tokens, while imposing light or weak distillation on tokens without a capacity advantage gap.
Extensive experiments across LLM and VLM settings demonstrate that DOPD consistently outperforms Vanilla OPD and strong competitive baselines, yielding superior distillation performance, robustness, continual-learning behavior, out-of-distribution generalization, and training stability.

\section{Limitations and Future Directions}
Notwithstanding the efficacy of our proposed DOPD, we acknowledge that several minor limitations remain.
First, our method hinges on the availability and quality of privileged information, the construction of which incurs additional costs for annotation, generation, and filtering processes.
Second, it introduces extra computational overhead relative to Vanilla OPD, requiring one additional forward pass of the student model.
Third, while the current routing strategy is intuitive, and empirically stable, it still relies on heuristic mechanisms.

Future research may further advance DOPD along directions: developing more reliable and cost-effective mechanisms for obtaining privileged information or discovering alternative strategy to detect available  advantage gap, with more principled or learnable distillation routing. More broadly, the paradigm of dynamic distillation from both teacher and student offers a useful lens for selective capacity transfer beyond LLMs and VLMs, inviting future work on more interpretable, efficient, and trustworthy distillation paradigms.

\bibliographystyle{plainnat}
\bibliography{cite}

\newpage
\beginappendix

\section{Details of Privileged Input}

\end{document}